\pdfoutput=1

\documentclass[11pt]{article}

\usepackage[final]{acl}

\usepackage{times}
\usepackage{latexsym}
\usepackage[T1]{fontenc}
\usepackage[utf8]{inputenc}
\usepackage{microtype}
\usepackage{inconsolata}
\usepackage{amssymb,mathtools}
\usepackage{algorithm}
\usepackage{algpseudocode}
\usepackage{dsfont}
\usepackage{graphicx,paralist,multirow}
\usepackage{makecell,booktabs}
\usepackage{cleveref}
\crefformat{section}{\S#2#1#3}
\crefformat{subsection}{\S#2#1#3}
\usepackage{colortbl,xcolor}
\usepackage{subcaption}
\usepackage{physics}
\usepackage{comment}

\newcommand{\Whole}{\textsc{Full}}
\newcommand{\OT}{\textsc{Oracle-T1}}
\newcommand{\OB}{\textsc{Oracle-B1}}
\newcommand{\Trans}{\textsc{Transfer-1}}
\newcommand{\TransO}{\textsc{Oracle-1}}
\newcommand{\Rand}{\textsc{Random}}
\newcommand{\CC}{\textsc{Calib}+}
\newcommand{\Comp}{\textsc{CompRW}}
\newcommand{\Vanilla}{\textsc{Standard}}
\newcommand{\PromptS}{\textsc{PromptS}}

\newcommand{\Hidden}{x}
\newcommand{\Attn}{a^{(l)}}
\newcommand{\MLP}{m^{(l)}}
\newcommand{\RR}{\mathbb{R}}
\newcommand{\AttnO}{W_{o}^{(l)}}
\newcommand{\Attnk}{W_{o_i}^{(l)}}
\newcommand{\HeadFinal}{\tilde{h}_i^{(l)}}
\newcommand{\Head}{h^{(l)}}
\newcommand{\CompLogit}{\vb{g}}
\newcommand{\Label}{\mathcal{Y}}
\newcommand{\Logits}{\mathcal{G}}
\newcommand{\Pred}{\mathcal{Z}}
\newcommand{\Prompt}{\emph{prompt}}
\newcommand{\ProbDist}{P_{rw}}
\newcommand{\Prob}{\vb{p}}
\newcommand{\WCC}{\vb{v}}

\newcommand{\Element}{z}
\newcommand{\InsideNorm}{\sum_j \Element_j}

\newcommand{\Ddev}{\mathcal{D}_\text{train}}
\newcommand{\Dtest}{\mathcal{D}_\text{test}}
\newcommand{\Ddemo}{\mathcal{D}_\text{demo}}

\title{When Parts Are Greater Than Sums:\\Individual LLM Components Can Outperform Full Models}

\author{
\textbf{Ting-Yun Chang}
\quad\quad \textbf{Jesse Thomason}
\quad\quad \textbf{Robin Jia} \\
University of Southern California, Los Angeles, CA, USA \\
\texttt{\{tingyun, jessetho, robinjia\}@usc.edu}
}

\begin{document}
\maketitle
\begin{abstract}
This paper studies in-context learning by decomposing the output of large language models into the individual contributions of attention heads and MLPs (\emph{components}).
We observe curious components: good-performing ones that individually do well on a classification task, even when the full model performs poorly; bad-performing ones that do much worse than chance; and label-biased components that always predict the same label.
We find that component accuracies are well-correlated across different demonstration sets and perturbations of prompt templates.
Based on our findings, we propose component reweighting, which learns to linearly re-scale the component activations from a few labeled examples.
Given $24$ labeled examples, our method improves by an average of $6.0\%$ accuracy points over $24$-shot ICL across 8 tasks on Llama-2-7B.
Overall, this paper both enriches our understanding of ICL and provides a practical method for improvement by examining model internals.

\end{abstract}

\section{Introduction}
\label{sec:intro}
The rapid progress in large language models (LLMs) has popularized prompting, which guides LLMs to perform tasks with instructions or examples.
Notably, in-context learning (ICL; \citealp{brown2020language}) adapts LLMs to a new task using only a few labeled examples without parameter updates.
However, how LLMs react to the in-context examples is sometimes unintuitive \cite{min-etal-2022-rethinking}. 
Recently, \citet{sclar2024quantifying} and \citet{voronov-etal-2024-mind} find that even for instruction-tuned \citep{ouyang2022training} or very large models, adding a space or newline in prompts can greatly affect accuracy.

\begin{figure}[t!]
  \centering
  \includegraphics[width=0.9\linewidth]{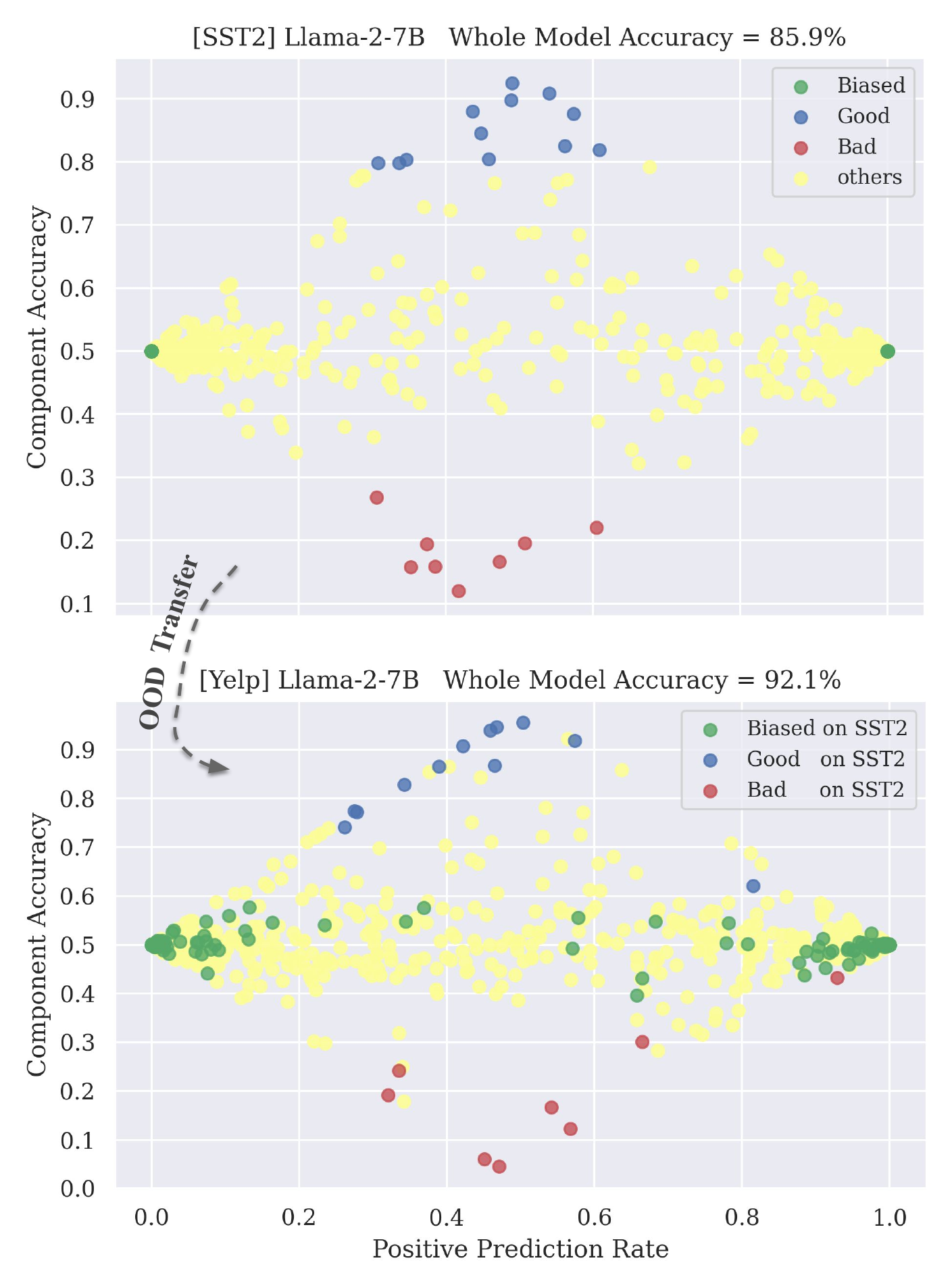}
      \caption{Each dot represents a component (attention head or MLP) under 4-shot ICL on Llama-2-7B. The $x$-axis shows how often a component predicts ``positive’’ on the test set. \textbf{Up:} We discover good-performing (blue), bad-performing (red), and label-biased (green) components. \textbf{Down:} Most components identified on SST2 show similar characteristics on Yelp-polarity.}
  \label{fig:daimond}
\end{figure}
We look into the LLM internals to understand what causes the surprising behavior across various ICL settings.
Our work stands in contrast to prior studies, which often treat LLMs as black boxes and alter either the input \cite{chen-etal-2023-relation, bertsch2024context} or output \cite{zhao2021calibrate,holtzman-etal-2021-surface}.
We introduce a new view of ICL by decomposing the output of an LLM into the sum of individual contributions of MLPs and attention heads, denoted ``components.''
Figure \ref{fig:daimond} reveals three types of curious components: good-performing ones (blue) that individually perform well or even outperform the full model, bad-performing ones (red) that perform below chance, and label-biased ones (green) that predict the same label on the entire test set.
We observe these three classes of components on Llama-2-7B, Llama-2-13B \cite{touvron2023llama}, Llama-3-8B \cite{dubey2024llama}, and Mistral-Instruct-7B \cite{jiang2023mistral} across 8 classification tasks.

We study the sensitivity of LLM components to multiple prompts formed by different demonstrations and templates.
We also construct contrast sets of templates---pairs of similar templates that yield large differences in ICL accuracy. 
Despite large variance in full-model accuracy, we find that component accuracies correlate well across different demonstrations ($r=0.80$ on average) and contrast set templates ($r=0.57$).
The top-performing components in contrast set pairs overlap and achieve decent accuracy even when the full model performs near random (Figure \ref{fig:main}).
Nonetheless, the component accuracies of two sampled templates are less correlated ($r=0.34$).
Further, good-performing components generalize well to out-of-distribution test sets.
For instance, the top-1 component for MNLI outperforms the full Llama-2-13B model by $9.1\%$ on MedNLI;
Figure \ref{fig:daimond} also shows that components are transferrable from SST2 to Yelp.
We conclude that components are relatively consistent in their behavior across prompts and datasets.

Inspired by our findings, we propose component reweighting.
Compared to prior work that selects prompts from a large pool of labeled data to improve ICL accuracy \cite{liu-etal-2022-makes}, component reweighting softly selects components by learning weights from few-shot examples to scale component activations.
Training these weights only involves learning a linear layer, which takes less than a minute on one CPU.
Overall, component reweighting better utilizes the same labeled examples, improving over $24$-shot ICL by $6.0\%, 2.2\%, 5.1\%, 1.6\%$ on Llama-2-7B, Llama-2-13B,  Mistral-Instruct-7B, and Llama-3-8B, respectively. At the same time, it enjoys similar inference speed as 4-shot ICL.

Finally, we study the training dynamics of components using the Pythia pretraining checkpoints \cite{biderman2023pythia}.
During pretraining, good-performing components emerge well before the full model performs well.
These findings suggest that LLMs acquire the internal ability to perform ICL early in training, but this ability only surfaces in the full model's behavior later on.

Overall, our work conducts extensive analysis of LLM internals, which motivates a practical method to improve ICL. 
We hope to inspire future work that further sheds light on LLM internals in order to improve performance. 
Our implementation is available at \url{https://github.com/terarachang/LLMDecomp}.
\section{Decomposing the Transformer in ICL}
\label{sec: decomp_method}
We introduce a new view of in-context learning by decomposing the Transformer architecture \cite{vaswani2017attention}. 
Our decomposition is exact---a mathematically equivalent formula for the model's outputs---and enables us to analyze model internals without training additional parameters (unlike, e.g., probing).
We first discuss what our new view offers over the standard view of ICL, and then walk through the mathematical details.

\subsection{A New View of In-Context Learning}
\begin{figure*}[t!]
  \centering
  \includegraphics[width=1.\linewidth]{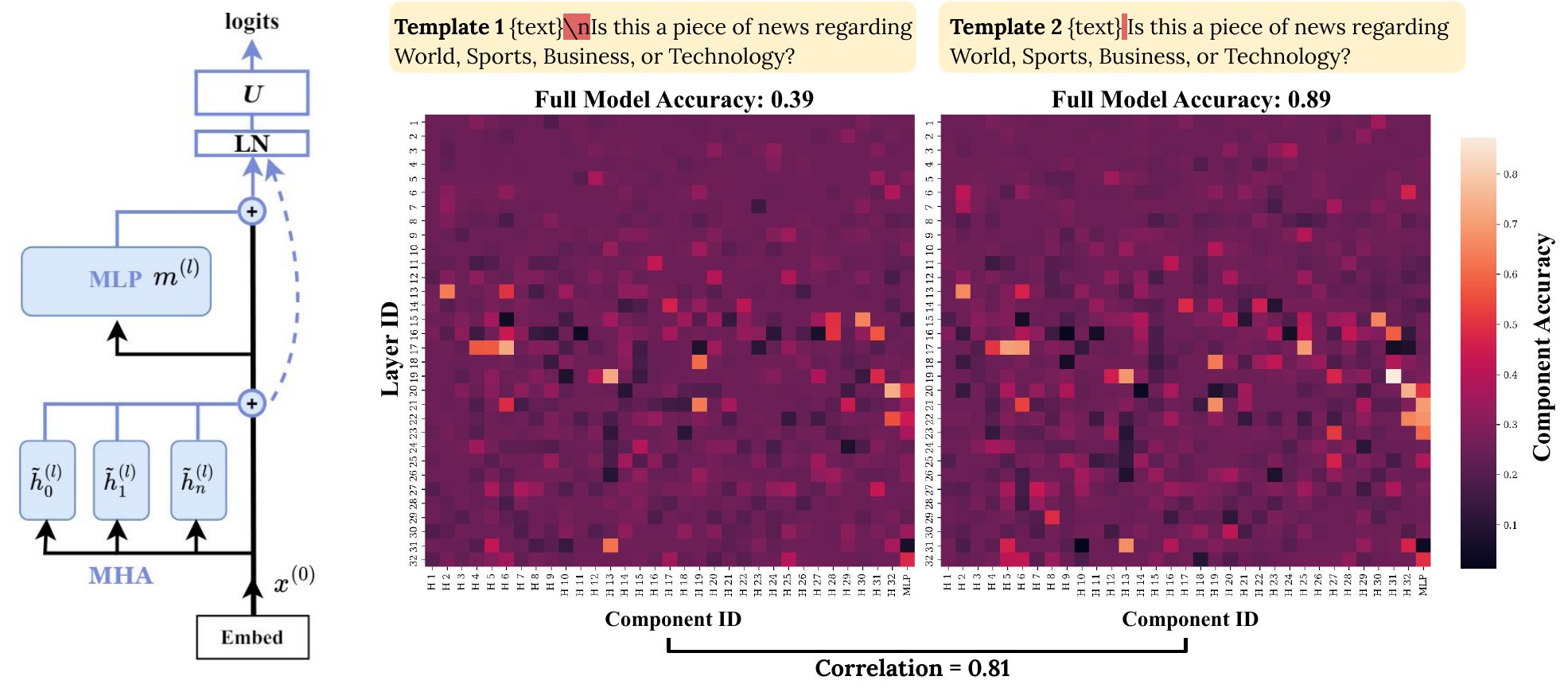}
      \caption{\textbf{Left:} Transformer decomposition. The components---MLPs and attention heads---are filled with blue, and the blue lines show the flow of early decoding. \textbf{Right:} We can calculate the individual accuracy of every component after decomposition. Although a pair of templates that only differ slightly yield very different accuracies ($0.39$ vs. $0.89$ on AGNews with Llama-2-7B), the accuracies of their internal components are highly correlated. The top components for Template 1 overlap with the ones for Template 2 and achieve $>0.7$ accuracy despite the poor full-model accuracy.}
  \label{fig:main}
\end{figure*}

\paragraph{Standard view.}
An LLM performs in-context learning (ICL) on a task based on a few demonstrations without training, where each demonstration is a templated example $(x, y)$ consisting of an input $x$ and a label word $y$.
We refer to a sequence of $K$ demonstrations $[x_1,y_1, \dotsc ,x_K,y_K]$ as a \Prompt.
The LLM makes predictions on a test input $x_{\text{test}}$ conditioned on the prompt, denoted by
${\arg\max}_{y\in \Label} P(y|\Prompt,x_{\text{test}})$, where $\Label$ is the set of possible label words in a classification task.
\paragraph{Our view.}
The residual stream of an LLM directly carries the information of the initial hidden state, every attention head, and every MLP, collectively named ``components,'' towards the output layer.
We view this information as the direct contributions\footnote{In comparison, a component has indirect contributions to the output by affecting other components in later layers \cite{wang2023interpretability}. This paper focuses on direction contributions.} of components to the output logits, and derive a formula for logits, $\sum_j \CompLogit_j$, where $\CompLogit_j$ is the direct contribution of the component indexed by $j$.
We can obtain the predictions of component $j$ with ${\arg\max}_{y\in \Label}\; \CompLogit_j$, and then calculate its individual ICL accuracy.
Specifically, we derive $\CompLogit_j = U \cdot C_j$ in Eq. \ref{eq:final_eq} below, where $U$ is the output embedding matrix and $C_j$ is the post-layernorm activations of component $j$.
We name the operation $(C_j \mapsto U \cdot C_j)$ as early decode, sharing the same spirit as \citet{nostalgebraist_interpreting_nodate} and \citet{geva-etal-2022-transformer}, which interpret hidden representations by decoding through $U$.
Compared to the standard view, we can directly study the behavior of individual components (Figure \ref{fig:main}), characterizing them and scaling their contributions to the model output.

\subsection{A Walkthrough of the Decomposition}
A Transformer of $L$ layers consists of a multi-headed attention (MHA) and MLP in every layer.
Let $\Attn \in \RR^{d}$ and $\MLP \in \RR^{d}$ be the output of the MHA and MLP at layer $l$, respectively. Due to residual connections, the hidden state $\Hidden^{(l)} \in \RR^{d}$ is: 
\begin{align}
    \Hidden^{(l)} &= \Hidden^{(l-1)} + \Attn + \MLP, \label{eq:residual} \\
    \Hidden^{(L)} &= \Hidden^{(0)} + \sum_{l=1}^{L} \; \left( \Attn + \MLP \right).
\end{align}
Note that GPT2-like LLMs apply layernorm before MHA and MLP \cite{radford2019language}; thus, layernorm is already taken into account as part of the formula for computing $\Attn$ and $\MLP$ (see \ref{sec:app_ln}).

An MHA $\Attn$ is composed of $n$ attention heads:
\begin{equation}
    \Attn = \AttnO \cdot \text{Concat}([\Head_1, \dots, \Head_n]) \label{eq:head}
\end{equation}
for $\Head_i \in \RR^{d_{\text{head}}}$ a head and $\AttnO \in \RR^{d \times n d_{\text{head}}}$ the output projection in MHA aggregating all heads.
\citet{elhage2021mathematical} rewrite Eq. \ref{eq:head} by segmenting $\AttnO$ into $n$ matrices $\Attnk\in \RR^{d \times d_{\text{head}}}$:
\begin{align}
    \Attn &= \sum_{i=1}^{n} \left( \Attnk \cdot \Head_i \right) = \sum_{i=1}^n \HeadFinal, \\
     & \text{where} \; [W_{o_1}^{(l)}, \dots, W_{o_n}^{(l)}] = \AttnO
\end{align}
Thus, we can treat each head as a single component adding $\HeadFinal = \Attnk \cdot \Head_i$ to the residual stream.

\begin{table*}[!t]
\begin{center}
\centering
\resizebox{1.8\columnwidth}{!}
{
\begin{tabular}{llcccccccccc}
\toprule

&& \textbf{SST2} & \textbf{BoolQ} & \textbf{QQP} & \textbf{WiC} & \textbf{RTE} & \textbf{MNLI} & \textbf{AGNews} & \textbf{ARC-Easy} & \textbf{Avg.} \\

\midrule
\multirow{3}{*}{\rotatebox[origin=c]{90}{\parbox[c]{1cm}{\centering \textbf{Llama2 7B}}}} 
&  \Whole & $75.8_{\hspace{0.05cm}18.1}$ & $69.2_{\hspace{0.05cm}12.0}$ & $61.3_{\hspace{0.05cm}9.9}$ & $52.4_{\hspace{0.05cm}3.0}$ & $\textbf{68.9}_{\hspace{0.05cm}3.2}$ & $34.4_{\hspace{0.05cm}1.7}$ & $70.0_{\hspace{0.05cm}19.9}$ & $\textbf{57.5}_{\hspace{0.05cm}14.4}$ & $61.2$\\
&  \OT & $\textbf{91.7}_{\hspace{0.05cm}\phantom{1}0.9}$ & $\textbf{69.7}_{\hspace{0.05cm}\phantom{1}7.7}$ & $\textbf{67.8}_{\hspace{0.05cm}4.3}$ & $\textbf{57.8}_{\hspace{0.05cm}1.1}$ & $64.6_{\hspace{0.05cm}2.7}$ & $\textbf{46.3}_{\hspace{0.05cm}3.3}$ & $\textbf{80.8}_{\hspace{0.05cm}\phantom{1}5.2}$ & $54.5_{\hspace{0.05cm}10.1}$ & $\textbf{66.6}$\\
&  \OB & $12.1_{\hspace{0.05cm}\phantom{1}2.7}$ & $34.1_{\hspace{0.05cm}\phantom{1}7.3}$ & $32.5_{\hspace{0.05cm}3.9}$ & $42.9_{\hspace{0.05cm}1.2}$ & $34.7_{\hspace{0.05cm}2.8}$ & $24.1_{\hspace{0.05cm}2.4}$ & $\phantom{1}3.0_{\hspace{0.05cm}\phantom{1}1.1}$ & $12.7_{\hspace{0.05cm}\phantom{1}4.2}$ & $24.5$\\
\midrule
\multirow{3}{*}{\rotatebox[origin=c]{90}{\parbox[c]{1cm}{\centering \textbf{Llama2 13B}}}} 
&   \Whole & $89.0_{\hspace{0.05cm}5.3}$ & $\textbf{77.6}_{\hspace{0.05cm}6.8}$ & $71.0_{\hspace{0.05cm}6.8}$ & $55.0_{\hspace{0.05cm}3.8}$ & $75.1_{\hspace{0.05cm}2.3}$ & $45.7_{\hspace{0.05cm}7.9}$ & $70.8_{\hspace{0.05cm}20.6}$ & $\textbf{73.2}_{\hspace{0.05cm}13.7}$ & $69.7$\\
&   \OT & $\textbf{92.5}_{\hspace{0.05cm}0.6}$ & $77.5_{\hspace{0.05cm}6.0}$ & $\textbf{73.5}_{\hspace{0.05cm}2.9}$ & $\textbf{60.4}_{\hspace{0.05cm}1.2}$ & $\textbf{75.7}_{\hspace{0.05cm}2.3}$ & $\textbf{56.4}_{\hspace{0.05cm}4.7}$ & $\textbf{84.6}_{\hspace{0.05cm}\phantom{1}3.6}$ & $73.1_{\hspace{0.05cm}\phantom{1}7.9}$ & $\textbf{74.2}$\\
&   \OB & $\phantom{1}8.2_{\hspace{0.05cm}1.0}$ & $27.1_{\hspace{0.05cm}9.7}$ & $31.8_{\hspace{0.05cm}3.4}$ & $39.5_{\hspace{0.05cm}1.6}$ & $27.9_{\hspace{0.05cm}2.8}$ & $18.6_{\hspace{0.05cm}2.6}$ & $\phantom{1}1.8_{\hspace{0.05cm}\phantom{1}0.9}$ & $\phantom{1}5.4_{\hspace{0.05cm}\phantom{1}3.5}$ & $20.0$\\
\midrule

\multirow{3}{*}{\rotatebox[origin=c]{90}{\parbox[c]{1.1cm}{\centering \textbf{Mistral Ins 7B}}}} 
&   \Whole & $90.1_{\hspace{0.05cm}2.9}$ & $\textbf{81.3}_{\hspace{0.05cm}2.1}$ & $70.9_{\hspace{0.05cm}7.2}$ & $58.5_{\hspace{0.05cm}4.2}$ & $80.5_{\hspace{0.05cm}1.7}$ & $56.1_{\hspace{0.05cm}5.0}$ & $83.0_{\hspace{0.05cm}5.7}$ & $\textbf{79.8}_{\hspace{0.05cm}1.4}$ & $75.0$\\
&   \OT & $\textbf{91.9}_{\hspace{0.05cm}0.7}$ & $80.8_{\hspace{0.05cm}2.0}$ & $\textbf{75.6}_{\hspace{0.05cm}2.6}$ & $\textbf{60.6}_{\hspace{0.05cm}2.2}$ & $\textbf{81.3}_{\hspace{0.05cm}0.8}$ & $\textbf{61.5}_{\hspace{0.05cm}3.3}$ & $\textbf{83.7}_{\hspace{0.05cm}4.3}$ & $78.5_{\hspace{0.05cm}2.2}$ & $\textbf{76.7}$\\
&   \OB & $\phantom{1}8.1_{\hspace{0.05cm}0.9}$ & $19.5_{\hspace{0.05cm}2.5}$ & $25.8_{\hspace{0.05cm}4.1}$ & $39.3_{\hspace{0.05cm}2.8}$ & $20.0_{\hspace{0.05cm}1.7}$ & $14.6_{\hspace{0.05cm}2.9}$ & $\phantom{1}1.8_{\hspace{0.05cm}0.7}$ & $\phantom{1}4.6_{\hspace{0.05cm}1.3}$ & $16.7$\\
\midrule

\multirow{3}{*}{\rotatebox[origin=c]{90}{\parbox[c]{1cm}{\centering \textbf{Llama3 8B}}}}
&   \Whole  & $91.4_{\hspace{0.05cm}1.7}$ & $\textbf{79.2}_{\hspace{0.05cm}7.2}$ & $74.0_{\hspace{0.05cm}8.0}$ & $58.7_{\hspace{0.05cm}4.7}$ & $\textbf{76.5}_{\hspace{0.05cm}2.2}$ & $59.4_{\hspace{0.05cm}3.7}$ & $\textbf{84.0}_{\hspace{0.05cm}6.6}$ & $\textbf{87.4}_{\hspace{0.05cm}5.5}$ & $76.3$\\
&   \OT & $\textbf{92.3}_{\hspace{0.05cm}1.0}$ & $77.4_{\hspace{0.05cm}7.3}$ & $\textbf{77.4}_{\hspace{0.05cm}3.7}$ & $\textbf{64.5}_{\hspace{0.05cm}2.7}$ & $76.3_{\hspace{0.05cm}2.9}$ & $\textbf{60.7}_{\hspace{0.05cm}1.5}$ & $81.4_{\hspace{0.05cm}5.7}$ & $86.0_{\hspace{0.05cm}5.9}$ & $\textbf{77.0}$\\
&   \OB & $\phantom{1}9.0_{\hspace{0.05cm}0.9}$ & $22.5_{\hspace{0.05cm}7.4}$ & $23.5_{\hspace{0.05cm}3.9}$ & $36.7_{\hspace{0.05cm}3.3}$ & $23.4_{\hspace{0.05cm}2.1}$ & $10.0_{\hspace{0.05cm}4.2}$ & $\phantom{1}1.4_{\hspace{0.05cm}0.6}$ & $\phantom{1}1.9_{\hspace{0.05cm}0.9}$ & $16.0$\\

\cmidrule(lr){0-10}
&   \Rand & $50.0$ & $50.0$ & $50.0$ & $50.0$ & $50.0$ & $33.3$ & $25.0$ & $25.0$ & $41.7$ \\
\bottomrule
\end{tabular}}

\end{center}

\caption{$\{3, 4\}$-shot ICL accuracy of 8 tasks and the average accuracy (\textbf{Avg.}). We run 15 prompts for each task (see \cref{sec:character}) and report the mean accuracy and standard deviation. We show the existence of good components (\OT) inside LLMs that individually perform on par with the full model (\Whole) on diverse tasks. Similarly, there exist bad components (\OB) that perform substantially below chance (\Rand). }
\label{table:top_bot}
\end{table*}

Finally, through the output embedding matrix $U \in \RR^{|\text{Vocab}| \times d}$, the output logits are:
\begin{align}
    & \text{logits} = U \cdot \text{LN} ( \Hidden^{(L)} ) \nonumber \\
    &= U \cdot \text{LN} \left( \Hidden^{(0)} + \sum_{l=1}^{L} \sum_{i=1}^{n} \HeadFinal + \sum_{l=1}^{L} \MLP \right) \nonumber \\
    &= U \cdot \text{LN} \left( \sum_{j=1}^{1+L\times n + L} \Element_j \right), \label{eq:flatten}
\end{align}
where $\Element = [\Hidden^{(0)}, \tilde{h}_1^{(1)}, \dotsc ,\tilde{h}_n^{(L)}, m^{(1)}, \dotsc , m^{(L)}]$ in Eq. \ref{eq:flatten} and we index every term in the summation with $j$.
LN$(\cdot)$ denotes the final layernorm, specifically, RMSNorm \cite{zhang2019root} for LLMs in our paper (see \ref{sec:app_ln}). 
In Eq. \ref{eq:flatten}, LN$(\InsideNorm) = \frac{\InsideNorm}{\text{RMS}(\InsideNorm)} \odot \gamma$, where RMS denotes root mean square, $\odot$ denotes element-wise multiplication, and $\gamma \in \RR^d$ is the affine parameters.
By pre-computing $\hat{\gamma} = \frac{\gamma}{\text{RMS}(\InsideNorm)}$, we have:
\begin{align}
    & \text{logits} = U \cdot \left( \InsideNorm \odot \hat{\gamma} \right) \\
    & = \sum_j U \cdot C_j, \; \text{where} \;C_j = \Element_j \odot \hat{\gamma} \label{eq:final_eq}
\end{align}
We refer to all $C_j \in \RR^{d}$ as the component activations, which include the activations of attention heads and MLPs after the final layernorm.\footnote{Empirically, we find that $\Hidden^{(0)}$ has near-random ICL accuracy on all the tasks, so we omit it in the rest of the paper.}
Now that we have broken down the Transformer output into simple additions in  Eq. \ref{eq:final_eq}, we can easily analyze the direct contribution of each component to the logits through the residual stream, $\CompLogit_j = U \cdot C_j$.

In ICL, we only need to do the decomposition when LLMs start to generate, i.e, when processing the last token of the input.
The computations on the other tokens are the same as the standard ICL.
In all our experiments, we use single-token label words. 
We use multiple templates from \citet{bach-etal-2022-promptsource} that cover diverse label words for each task.
\section{Characterizing Components for ICL}
\label{sec:character}
We conduct in-context learning across 8 classification tasks on 4 LLMs: Llama-2-7B, Llama-2-13B, Mistral-Instruct-7B, and Llama-3-8B.
ICL is sensitive to prompts, so we randomly sample 5 disjoint sets of demonstrations formatted with 3 templates and report the standard deviation across the 15 runs.
To avoid majority and recency biases \cite{zhao2021calibrate}, each prompt consists of the same number of demonstrations from every class in shuffled order.
We use $K=3$ demonstrations for 3-way classification tasks and $K=4$ for the other tasks. 
Except for \cref{sec:reweight}, we refer to $K=\{3, 4\}$ without further notice.
We sample 2000 examples with balanced labels as the test set for every task.
Please see \ref{sec:task_details} for details about the tasks and templates.

\subsection{Good and Bad-Performing Components}
\label{sec:top}
Across all the tasks and LLMs, we observe good-performing components that perform well or even outperform the full model, and bad-performing components that individually perform much worse than chance (blue and red dots in Figure~\ref{fig:daimond}, respectively).
Table \ref{table:top_bot} compares the full model (\Whole) with the top-1 (\OT) and bottom-1 (\OB) components selected on the test set.
On average, \OT\ outperforms \Whole\ by $5.4\%, 4.5\%, 1.7\%, 0.7\%$ on Llama-2-7B, Llama-2-13B, Mistral-Instruct-7B, and Llama-3-8B, respectively; \OB\ underperforms random guessing (\Rand) by $17.2\%, 20.7\%, 25.0\%, 25.7\%$. 

\subsection{Label-Biased Components}
\label{sec:bias}
Besides good and bad-performing components, we also observe label-biased components, which predict a certain label on the entire test set (the green dots in Figure \ref{fig:daimond}).
These components exist in all the tasks and LLMs we study, accounting for $29.1\%, 26.4\%, 22.8\%, 29.7\%$ of components on average in Llama-2-7B, Llama-2-13B, Mistral-Instruct-7B, and Llama-3-8B, respectively (Table \ref{table:label_bias}).
In \ref{sec:app_biased}, we show that even when we prompt the model with all demonstrations of positive labels, the most biased component still insists on predicting ``negative'' on the entire test set, and vice versa.

\subsection{Mechanistic Understanding of Bad-Performing Heads}
\label{sec:mech}
Prior work studies the mechanism of certain components in LLMs, showing that
there are negative mover attention heads that write in the opposite direction of the expected answer \cite{wang2023interpretability} and copy suppression heads that suppress the prediction of a prior token in the context \cite{mcdougall2023copy}.
Inspired by them, we investigate the mechanism behind bad-performing heads identified by our decomposition.
We focus on label tokens in the context, as \citet{wang-etal-2023-label} show that label words serve as anchors in ICL.

We conduct a case study on Llama-2-7B with 4-shot balanced in-context examples from SST2.
We examine the bottom-5 attention heads that have the worst ICL accuracy on SST2.
We find that three of these heads, L19H15, L15H14, and L18H9, assign top attention probabilities to all 4 label tokens of the 4-shot in-context examples when predicting test examples. 
Furthermore, despite their poor ICL accuracy, these heads actually assign higher attention to the correct in-context label tokens than the incorrect ones most of the time ($> 70\%$ of the test examples).
In other words, when a test example has a positive label, these heads assign higher attention\footnote{We average the attention probabilities of the same label tokens and then compare the average ones of the two labels.} to the tokens ``positive’’ in the context than the tokens ``negative’’.
We also observe that the more the heads attend to ``positive’’ in the context, the lower the inner product between the head and the output embedding of the token ``positive’’, with the correlation $r=-0.97, -0.96, -0.89$ for L19H15, L15H14, and L18H9, respectively. 

 In summary, we show that some bad-performing heads attend highly to prior label tokens and decrease the output probability of the correct one, which shares similarities with the copy suppression heads and negative mover heads \cite{mcdougall2023copy,wang2023interpretability}.
 However, we do not observe similar behavior in other tasks, where the bad-performing heads usually attend to ``<s>’’, ``?’’, or ``\textbackslash n’’.
 We invite future work to further analyze how bad-performing heads function in general.
\begin{table}[!t]
\begin{center}
\centering

\resizebox{1.\columnwidth}{!}
{
\begin{tabular}{llccccc}
\toprule

&& \textbf{SST2} & \textbf{BoolQ} & \textbf{QQP} & \textbf{AGNews} & \textbf{ARC} \\
\midrule
\multirow{3}{*}{\rotatebox[origin=c]{90}{\parbox[c]{1cm}{\centering \textbf{Corr}}}} 
& (1) Demo & 0.81 & 0.84 & 0.60 & 0.89 & 0.88 \\
& (2) Temp &0.40 & 0.16 & 0.03 & 0.68 & 0.44 \\
& (3) Cst T&0.72 & 0.63 & 0.23 & 0.82 & 0.46 \\
\midrule
\multirow{3}{*}{\rotatebox[origin=c]{90}{\parbox[c]{1cm}{\centering \textbf{IoU}}}} 
& (1) Demo & 0.36 & 0.74 & 0.27 & 0.63 & 0.70 \\
& (2) Temp & 0.12 & 0.01 & 0.01 & 0.20 & 0.20 \\
& (3) Cst T& 0.40 & 0.23 & 0.02 & 0.36 & 0.45 \\
\bottomrule
\end{tabular}}

\end{center}

\caption{The average correlation and IoU between (1) two random sets of demonstrations, (2) two random templates, and (3) two minimally contrastive templates.}
\label{table:iou}
\end{table}

\section{Transferability of Components}
\label{sec:transfer}
We observe moderate to high component transferability across demonstrations, minimally contrastive templates, and data distributions, whereas there is little transferability across randomly sampled templates.
Our decomposition uncovers hidden abilities of individual components when the full model performs poorly.

\subsection{Transfer across Prompt Variants}
\label{sec:demo_template}
We first measure the agreement in component accuracies between (1) two disjoint sets of demonstrations with a fixed template, (2) two randomly sampled templates with fixed demonstrations, and (3) two minimally-contrastive templates with fixed demonstrations.
Recall that we have 5 sets of demonstrations and 3 templates in total (\cref{sec:character}); here, we calculate the average agreement between every pair.
For (3), we construct contrast sets \cite{gardner-etal-2020-evaluating} by minimally editing the worst-performing template out of the 3 templates into a good template, which yields at least $10\%$ improvement in average accuracy.
Our edits include adding a space, removing a newline, or changing label words (see Table \ref{table:appendix_cst}).
We use two metrics to measure the agreement between each pair: Pearson correlation of the accuracies of all components and the intersection over union (IoU) on the sets of top-5 components, which measures whether the top-performing components of the pair overlap.

Table \ref{table:iou} summarizes the results on Llama-2-7B;
\ref{sec:full_transfer} shows similar findings on other models.
(1) The accuracies of the internal components are highly consistent across different choices of demonstrations, having strong correlations and an average of $0.54$ IoU.
(2) The components have much weaker agreement across randomly sampled templates, having a near 0 IoU on BoolQ and QQP.
(3) Nevertheless, there is agreement between minimally contrastive templates (Cst T), with an average correlation of $0.57$ across tasks, despite contrasting full-model accuracy. 
For example, Figure \ref{fig:main} demonstrates that full-model accuracy changes dramatically ($39\%$ vs $89\%$) in a minimal pair of templates, but internal components have a high correlation of $0.81$ and the pair shares top-performing components.
Combining (2) and (3) suggests components behave similarly on similar templates, but this similarity decreases as the templates diverge.

\subsection{Transfer to Out-of-Distribution Test Sets}
\label{sec:ood}
We further study whether the best component selected on the test set can still perform well on an out-of-distribution (OOD) test set.
We name this method, which uses a single component to make predictions, as \Trans.
Specifically, we study component transferability from SST2 to Yelp-polarity, MNLI to MedNLI, and BoolQ to BoolQ Contrast Set.
We compare \Trans\ with using the full model (\Whole) on the OOD test sets.
To understand the best possible \Trans{} accuracy, we also report the best component accuracy directly selected on the OOD set, \TransO.

Table \ref{table:ood} shows that \Trans\ closely matches \TransO\ overall, suggesting that the top-performing components are transferable across data distribution.
Moreover, \Trans\ sometimes outperforms \Whole, especially on Llama2 models, showing the hidden abilities of the internal components.

\subsection{Transfer between Two Opposite Tasks}
\label{sec:task69}
We conduct a case study of component transferability across instructions using Task069 and Task070 of Super-NaturalInstructions \cite{wang-etal-2022-super}, both of which are binary abductive NLI tasks \cite{Bhagavatula2020Abductive}.
The instruction for Task069 asks for correct answers, while Task070 asks for incorrect ones (``pick the one that makes less sense;’’ see Figure~\ref{fig:templates_69_70} for the full instructions).
Examples in the two tasks are not parallel.

We find that Mistral-Instruct-7B achieves good accuracy across 15 runs on Task069 ($76.8 \pm 2.4$), but below chance on Task070 ($40.6 \pm 5.4$).
We observe a strong negative correlation, $r = -0.60$ on average, between the component accuracies of the two tasks.
The worst-performing components in Task069 become the top-performing in Task070 and vice versa.
The correlation suggests that the model has the ability to solve Task070, but misunderstands negation.
Thus, we apply the \mbox{\Trans} method (\cref{sec:ood}) but select the worst-performing component from Task069 and then calculate its individual accuracy on Task070.
\mbox{\Trans} achieves $58.7 \pm 4.8$ accuracy across the 15 runs, an \textbf{improvement of} $\mathbf{18.1\%}$ over the full model.
These results suggest that components behave consistently even across tasks with opposite instructions, as the active components in Task069 are also active in Task070.
\begin{table}[!t]
\begin{center}
\centering
\resizebox{1.\columnwidth}{!}
{
\begin{tabular}{llccc}
\toprule
&& \textbf{Yelp-polarity} & \textbf{MedNLI} & \textbf{BoolQ Cst} \\
\midrule
\multirow{3}{*}{\rotatebox[origin=c]{90}{\parbox[c]{1cm}{\centering \textbf{Llama2 7B}}}} 
&  \Whole & $84.7_{\hspace{0.05cm}15.4}$ & $34.3_{\hspace{0.05cm}1.7}$ & $\textbf{64.9}_{\hspace{0.05cm}9.8}$\\
&  \Trans & $\textbf{94.9}_{\hspace{0.05cm}\phantom{1}3.1}$ & $\textbf{42.6}_{\hspace{0.05cm}4.7}$ & $64.3_{\hspace{0.05cm}7.9}$\\
&  \TransO & $96.9_{\hspace{0.05cm}\phantom{1}0.7}$ & $48.8_{\hspace{0.05cm}2.3}$ & $66.2_{\hspace{0.05cm}5.7}$\\
\midrule
\multirow{3}{*}{\rotatebox[origin=c]{90}{\parbox[c]{1cm}{\centering \textbf{Llama2 13B}}}} 
&  \Whole & $95.9_{\hspace{0.05cm}1.4}$ & $46.8_{\hspace{0.05cm}9.6}$ & $72.0_{\hspace{0.05cm}7.6}$\\
&  \Trans & $\textbf{96.0}_{\hspace{0.05cm}1.8}$ & $\textbf{55.9}_{\hspace{0.05cm}4.0}$ & $\textbf{72.3}_{\hspace{0.05cm}6.5}$\\
&  \TransO & $97.1_{\hspace{0.05cm}0.4}$ & $57.0_{\hspace{0.05cm}3.7}$ & $73.0_{\hspace{0.05cm}6.1}$\\
\midrule

\multirow{3}{*}{\rotatebox[origin=c]{90}{\parbox[c]{1.1cm}{\centering \textbf{Mistral 
 Ins 7B}}}} 
&  \Whole & $\textbf{97.0}_{\hspace{0.05cm}0.5}$ & $57.3_{\hspace{0.05cm}5.7}$ & $\textbf{74.6}_{\hspace{0.05cm}3.5}$\\
&  \Trans & $95.6_{\hspace{0.05cm}1.6}$ & $\textbf{61.9}_{\hspace{0.05cm}4.8}$ & $73.7_{\hspace{0.05cm}3.7}$\\
&  \TransO & $97.1_{\hspace{0.05cm}0.4}$ & $62.7_{\hspace{0.05cm}4.1}$ & $74.5_{\hspace{0.05cm}3.6}$\\
\midrule
\multirow{3}{*}{\rotatebox[origin=c]{90}{\parbox[c]{1cm}{\centering \textbf{Llama3 
 8B}}}}  
 & \Whole & $\textbf{97.8}_{\hspace{0.05cm}0.4}$ & $61.0_{\hspace{0.05cm}2.2}$ & $\textbf{77.3}_{\hspace{0.05cm}7.5}$\\
&  \Trans & $95.9_{\hspace{0.05cm}4.4}$ & $\textbf{61.3}_{\hspace{0.05cm}0.8}$ & $73.9_{\hspace{0.05cm}8.4}$\\
& \TransO & $97.9_{\hspace{0.05cm}0.5}$ & $61.6_{\hspace{0.05cm}0.6}$ & $74.8_{\hspace{0.05cm}8.9}$\\

\bottomrule

\end{tabular}}

\end{center}

\caption{The average ICL accuracy and standard deviation on OOD test sets. The components selected on the in-distribution test sets (\Trans) can transfer to OOD sets, performing similarly to the oracle components (\TransO) directly selected on the OOD sets.}
\label{table:ood}
\end{table}
\section{Component Reweighting}
\subsection{Proposed Method}
\label{sec:reweight}
Our findings in \cref{sec:transfer} show the promising direction of selecting internal components to improve ICL.
Therefore, we propose a method that reweights components by learning a weight $w_j \in \RR$ on every component activation $C_j$. 
Reweighting is a soft version of selection, which can be learned by gradient descent on very few examples.

Given $K$ labeled examples, instead of using all of them as ICL demonstrations, we divide them into a demonstration set $\Ddemo$ and a training set $\Ddev$. 
We first randomly sample $K^\prime=\{3,4\}$ examples with balanced labels as demonstrations and use the remaining examples as $\Ddev$ to train the component weights.
Specifically, we can rewrite Eq. \ref{eq:final_eq} as $\text{logits} = \sum_j w_j (U \cdot C_j)$, where $w_j = 1$ for all $j$.
Because of the existence of good and bad-performing components, weighing all components equally may not be optimal. 
Therefore, we tune the weights $w \in \RR^N$ of $N$ components on $\Ddev$ with cross-entropy loss and $L_1$ regularization, while keeping the LLM frozen:
\begin{align}
&\mathcal{L} = \sum_{(x,y)\in \Ddev} -\log \ProbDist(y | x) \label{eq:reweight} + \lambda \|w\|_1, \\
&\ProbDist(y | x) = \text{softmax}\left(\sum_{j=1}^{N} w_j\;(U_{\Label} \cdot C_j)\right)_y, \nonumber
\end{align}
where $U_{\Label} \in \RR^{|\Label| \times d}$ is a submatrix of $U$ that comprises the output embeddings of label words, $\ProbDist$ is the probability distribution of the LLM after reweighting, and $\lambda$ is the hyperparameter of the $L_1$ loss to encourage sparsity on the component weights.
We obtain the activations $\{C_j\}_{j=1}^N$ of all components in one $K^\prime$-shot forward pass, computed on the prompt derived from $\Ddemo$, followed by $x$.
Our method scales each component's direct contributions to the logits ($U_{\Label} \cdot C_j \in \RR^{|\Label|}$) by $w_j$.
In practice, we cache these contributions on the entire training set as input features to the linear layer $w$, which allows us to discard the entire LLM while training $w$ (line 9 and 13 in Algorithm \ref{algo:comp}), saving tremendous training time and GPU memory.
The cache only requires $O(|\Label| \times N \times |\Ddev|)$ space.
At inference time, the overhead of our method over $K^\prime$-shot ICL is to early decode $N$ components and apply the learned weights, i.e., $\sum_{j=1}^N w_j\;(U_{\Label} \cdot C_j)$.
As both $|\Label|$ and $N$ are small ($N < 2000$ for all LLMs in this paper), the overhead is negligible compared to the computation of the LLM itself.

\begin{algorithm}[!t]
\small
  \caption{Component Reweighting}
  \label{algo:comp}
\begin{algorithmic}[1]
\State {\bfseries Input:} $K$ labeled examples, a test set $\Dtest$, a set of label words $\Label$, an LLM $\mathcal{M}$, the number of components $N$
\State {\bfseries Output:} $\Pred$, the predictions of $\mathcal{M}$ on $\Dtest$

\State Split $K$ examples into a \Prompt\ consists of $K^\prime$ demonstrations and a training set $\Ddev$ of $K-K^\prime$ examples
\State $U_{\Label} \gets$ concatenate the output embeddings of $\mathcal{Y}$ in $\mathcal{M}$

\State Initialize $\Logits^{\text{train}} \gets \varnothing$
\For { $(x, y) \in \Ddev$ }
\State $\{C_j\}_{j=1}^{N} \gets$ $\mathcal{M}(\Prompt, x)$ \textcolor{gray}{\Comment{$K^\prime$-shot ICL}}

\For{ $j \gets 1$ to $N$ }
\State$\Logits^{\text{train}} \gets \Logits^{\text{train}} \cup (U_{\Label} \cdot C_j)$ \textcolor{gray}{\Comment{early decode}}
\EndFor
\EndFor
\State Initialize $w \gets [1, \dotsc, 1] \in \RR^{N}$
\State Train the weights $w$ on $\Logits^{\text{train}}$ with Eq. \ref{eq:reweight}

\State Initialize $\Pred\ \gets \varnothing$ \textcolor{gray}{\Comment{Start Inference}}
\For { $(x, y) \in \Dtest$ } 
\State $\{C_j\}_{j=1}^{N} \gets$ $\mathcal{M}(\Prompt, x)$ \textcolor{gray}{\Comment{$K^\prime$-shot ICL}}

\State Initialize $\CompLogit \gets [0, \dotsc, 0] \in \RR^{|\Label|}$ 
\For{ $j \gets 1$ to $N$ } \textcolor{blue}{\Comment{Test-Time Overhead}}
\State $\CompLogit \gets \CompLogit + w_j(U_{\Label} \cdot C_j)$ \textcolor{gray}{\Comment{early decode}}
\EndFor
\State $\hat{y} \gets {\arg\max}_{y\in \Label}\; \CompLogit$
\State $\Pred\ \gets \Pred\ \cup \hat{y}$ 
\EndFor

\State \textbf{return} $\Pred$
\end{algorithmic}
\end{algorithm}
\subsection{Baselines}
\label{sec:baseline}

\paragraph{Standard ICL.}
The simplest baseline is to use all the $K$ labeled examples as demonstrations.
Since the other methods use $K^\prime$ examples as demonstrations, we report the accuracy of standard \mbox{$K^\prime$-shot} ICL using the same $\Ddemo$ for reference.

\paragraph{Prompt Selection.}
\citet{liu-etal-2022-makes} improve ICL accuracy by selecting demonstrations from a pool of labeled data for each test example.
Here, we select from the given $K$ labeled examples.
Following \citet{rubin-etal-2022-learning}, we use SBERT \cite{reimers-gurevych-2019-sentence} to encode examples into sentence embeddings and select the $K^\prime=\{3, 4\}$ nearest neighbors under cosine similarity as the demonstrations for each test example.

\paragraph{Calibration.}
As LLMs tend to predict a certain class over others, \citet{zhao2021calibrate} reweight the output class probabilities.
They use context-free inputs, such as ``N/A’’, to calibrate the probability distribution.
However, \citet{fei-etal-2023-mitigating} and \citet{zhou2023batch} find context-free inputs sometimes ineffective, because in-domain context is important for calibration.
Thus, we introduce \CC, which calibrates the original probabilities $\Prob \in \RR^{|\Label|}$ with a training set of in-distribution labeled examples, $\Ddev$.
We train the calibration weights $\WCC \in \RR^{|\Label|}$ on $\Ddev$ with cross-entropy loss and obtain the calibrated probabilities $\hat{\Prob} = \text{softmax}(\WCC \cdot \Prob)$.
For direct comparisons, we split the $K$ examples into the same $\Ddemo$ and $\Ddev$ sets as component reweighting for \CC, where $|\Ddemo| = K^\prime$.
We include the training details of both methods in \ref{sec:app_hyper}.

\begin{table*}[!t]
\begin{center}
\centering
\resizebox{1.85\columnwidth}{!}
{
\begin{tabular}{llcccccccccc}
\toprule

&& \textbf{SST2} & \textbf{BoolQ} & \textbf{QQP} & \textbf{WiC} & \textbf{RTE} & \textbf{MNLI} & \textbf{AGNews} & \textbf{ARC-Easy} & \textbf{Avg.} \\

\cmidrule(lr){2-11}
\parbox[t]{2mm}{\multirow{9}{*}{\rotatebox[origin=c]{90}{\textbf{Llama-2-7B}}}}
&~ \Vanilla\ $3,4$ & $75.8_{\hspace{0.05cm}18.1}$ & $69.2_{\hspace{0.05cm}12.0}$ & $61.3_{\hspace{0.05cm}9.9}$ & $52.4_{\hspace{0.05cm}3.0}$ & $68.9_{\hspace{0.05cm}3.2}$ & $34.4_{\hspace{0.05cm}1.7}$ & $70.0_{\hspace{0.05cm}19.9}$ & $57.5_{\hspace{0.05cm}14.4}$ & $61.2$\\
\cmidrule(lr){2-11}
&~  \Vanilla\ $12$ & $77.8_{\hspace{0.05cm}19.6}$ & $\textbf{71.6}_{\hspace{0.05cm}\phantom{1}8.0}$ & $63.6_{\hspace{0.05cm}7.8}$ & $52.5_{\hspace{0.05cm}2.4}$ & $\textbf{71.1}_{\hspace{0.05cm}2.1}$ & $37.0_{\hspace{0.05cm}2.8}$ & $69.0_{\hspace{0.05cm}20.8}$ & $\textbf{59.6}_{\hspace{0.05cm}13.9}$ & $62.8$\\
&~  \PromptS\ $12$ & $73.8_{\hspace{0.05cm}19.2}$ & $69.4_{\hspace{0.05cm}10.5}$ & $62.2_{\hspace{0.05cm}6.1}$ & $53.1_{\hspace{0.05cm}2.7}$ & $65.5_{\hspace{0.05cm}1.8}$ & $35.5_{\hspace{0.05cm}1.6}$ & $59.1_{\hspace{0.05cm}28.7}$ & $58.7_{\hspace{0.05cm}11.9}$ & $59.7$\\
&~  \CC\ $12$ & $85.1_{\hspace{0.05cm}\phantom{1}6.0}$ & $69.2_{\hspace{0.05cm}13.6}$ & $\textbf{73.6}_{\hspace{0.05cm}6.1}$ & $55.1_{\hspace{0.05cm}5.1}$ & $70.3_{\hspace{0.05cm}2.7}$ & $45.5_{\hspace{0.05cm}7.8}$ & $77.8_{\hspace{0.05cm}12.2}$ & $58.6_{\hspace{0.05cm}14.6}$ & $66.9$\\
&~  \Comp\ $12$ & $\textbf{88.5}_{\hspace{0.05cm}\phantom{1}2.8}$ & $70.4_{\hspace{0.05cm}11.2}$ & $71.4_{\hspace{0.05cm}5.4}$ & $\textbf{56.3}_{\hspace{0.05cm}3.4}$ & $70.0_{\hspace{0.05cm}2.8}$ & $\textbf{48.3}_{\hspace{0.05cm}4.8}$ & $\textbf{87.4}_{\hspace{0.05cm}\phantom{1}2.3}$ & $58.3_{\hspace{0.05cm}13.6}$ & $\textbf{68.8}$\\
\cmidrule(lr){2-11}
&~  \Vanilla\ $24$ & $77.8_{\hspace{0.05cm}19.5}$ & $71.6_{\hspace{0.05cm}\phantom{1}7.3}$ & $66.4_{\hspace{0.05cm}5.0}$ & $53.2_{\hspace{0.05cm}3.3}$ & $\textbf{71.9}_{\hspace{0.05cm}1.5}$ & $39.9_{\hspace{0.05cm}3.6}$ & $71.1_{\hspace{0.05cm}20.0}$ & $58.3_{\hspace{0.05cm}16.2}$ & $63.8$\\
&~  \PromptS\ $24$ & $74.2_{\hspace{0.05cm}20.4}$ & $68.9_{\hspace{0.05cm}10.2}$ & $62.1_{\hspace{0.05cm}4.9}$ & $53.6_{\hspace{0.05cm}1.9}$ & $64.8_{\hspace{0.05cm}0.9}$ & $36.4_{\hspace{0.05cm}1.5}$ & $57.5_{\hspace{0.05cm}30.2}$ & $58.0_{\hspace{0.05cm}12.5}$ & $59.4$\\
&~  \CC\ $24$ & $87.6_{\hspace{0.05cm}\phantom{1}5.0}$ & $70.3_{\hspace{0.05cm}11.9}$ & $\textbf{73.4}_{\hspace{0.05cm}5.5}$ & $55.8_{\hspace{0.05cm}4.9}$ & $70.4_{\hspace{0.05cm}2.7}$ & $46.4_{\hspace{0.05cm}6.7}$ & $78.4_{\hspace{0.05cm}11.8}$ & $\textbf{59.2}_{\hspace{0.05cm}14.4}$ & $67.7$\\
&~  \Comp\ $24$ & $\textbf{90.6}_{\hspace{0.05cm}\phantom{1}1.7}$ & $\textbf{71.7}_{\hspace{0.05cm}\phantom{1}9.4}$ & $71.9_{\hspace{0.05cm}4.4}$ & $\textbf{57.1}_{\hspace{0.05cm}3.0}$ & $70.0_{\hspace{0.05cm}4.1}$ & $\textbf{49.8}_{\hspace{0.05cm}4.0}$ & $\textbf{88.1}_{\hspace{0.05cm}\phantom{1}2.1}$ & $58.8_{\hspace{0.05cm}13.6}$ & $\textbf{69.8}$\\
\midrule
\parbox[t]{2mm}{\multirow{9}{*}{\rotatebox[origin=c]{90}{\textbf{Llama-2-13B}}}}
&~  \Vanilla\ $3,4$ & $89.0_{\hspace{0.05cm}\phantom{1}5.3}$ & $77.6_{\hspace{0.05cm}6.8}$ & $71.0_{\hspace{0.05cm}6.8}$ & $55.0_{\hspace{0.05cm}3.8}$ & $75.1_{\hspace{0.05cm}2.3}$ & $45.7_{\hspace{0.05cm}7.9}$ & $70.8_{\hspace{0.05cm}20.6}$ & $73.2_{\hspace{0.05cm}13.7}$ & $69.7$\\
\cmidrule(lr){2-11}
&~  \Vanilla\ $12$ & $\textbf{91.3}_{\hspace{0.05cm}\phantom{1}1.9}$ & $78.1_{\hspace{0.05cm}7.4}$ & $70.5_{\hspace{0.05cm}7.3}$ & $\textbf{59.6}_{\hspace{0.05cm}2.4}$ & $74.4_{\hspace{0.05cm}3.5}$ & $55.1_{\hspace{0.05cm}6.2}$ & $84.7_{\hspace{0.05cm}\phantom{1}7.8}$ & $71.2_{\hspace{0.05cm}16.4}$ & $73.1$\\
&~  \PromptS\ $12$ & $83.8_{\hspace{0.05cm}10.2}$ & $74.9_{\hspace{0.05cm}6.6}$ & $64.6_{\hspace{0.05cm}5.7}$ & $57.0_{\hspace{0.05cm}2.1}$ & $69.5_{\hspace{0.05cm}3.5}$ & $48.1_{\hspace{0.05cm}5.4}$ & $64.4_{\hspace{0.05cm}29.6}$ & $74.2_{\hspace{0.05cm}\phantom{1}9.3}$ & $67.1$\\
&~  \CC\ $12$ & $89.4_{\hspace{0.05cm}\phantom{1}3.2}$ & $\textbf{78.4}_{\hspace{0.05cm}6.1}$ & $72.1_{\hspace{0.05cm}4.1}$ & $58.1_{\hspace{0.05cm}5.1}$ & $75.3_{\hspace{0.05cm}1.9}$ & $57.3_{\hspace{0.05cm}4.5}$ & $81.5_{\hspace{0.05cm}\phantom{1}8.7}$ & $74.7_{\hspace{0.05cm}\phantom{1}9.3}$ & $73.3$\\
&~  \Comp\ $12$ & $89.1_{\hspace{0.05cm}\phantom{1}3.2}$ & $77.7_{\hspace{0.05cm}6.7}$ & $\textbf{72.7}_{\hspace{0.05cm}3.3}$ & $58.7_{\hspace{0.05cm}4.0}$ & $\textbf{76.2}_{\hspace{0.05cm}2.0}$ & $\textbf{60.2}_{\hspace{0.05cm}3.7}$ & $\textbf{88.1}_{\hspace{0.05cm}\phantom{1}1.7}$ & $\textbf{76.2}_{\hspace{0.05cm}\phantom{1}6.8}$ & $\textbf{74.9}$\\
\cmidrule(lr){2-11}
&~  \Vanilla\ $24$ & $\textbf{91.9}_{\hspace{0.05cm}\phantom{1}0.6}$ & $77.7_{\hspace{0.05cm}8.2}$ & $69.5_{\hspace{0.05cm}8.5}$ & $60.6_{\hspace{0.05cm}1.6}$ & $74.7_{\hspace{0.05cm}3.3}$ & $58.2_{\hspace{0.05cm}7.0}$ & $85.8_{\hspace{0.05cm}\phantom{1}4.4}$ & $69.1_{\hspace{0.05cm}17.7}$ & $73.5$\\
&~  \PromptS\ $24$ & $81.9_{\hspace{0.05cm}13.2}$ & $75.1_{\hspace{0.05cm}5.7}$ & $64.9_{\hspace{0.05cm}4.8}$ & $57.3_{\hspace{0.05cm}1.8}$ & $69.5_{\hspace{0.05cm}1.7}$ & $49.8_{\hspace{0.05cm}5.1}$ & $65.2_{\hspace{0.05cm}28.9}$ & $74.2_{\hspace{0.05cm}\phantom{1}9.4}$ & $67.2$\\
&~  \CC\ $24$ & $90.7_{\hspace{0.05cm}\phantom{1}2.1}$ & $\textbf{78.6}_{\hspace{0.05cm}6.2}$ & $73.1_{\hspace{0.05cm}4.3}$ & $\textbf{59.5}_{\hspace{0.05cm}3.2}$ & $75.9_{\hspace{0.05cm}1.9}$ & $58.4_{\hspace{0.05cm}2.8}$ & $82.0_{\hspace{0.05cm}\phantom{1}8.4}$ & $75.2_{\hspace{0.05cm}\phantom{1}9.1}$ & $74.2$\\
&~  \Comp\ $24$ & $91.0_{\hspace{0.05cm}\phantom{1}1.8}$ & $78.2_{\hspace{0.05cm}6.4}$ & $\textbf{74.2}_{\hspace{0.05cm}3.1}$ & $58.5_{\hspace{0.05cm}4.1}$ & $\textbf{77.1}_{\hspace{0.05cm}1.8}$ & $\textbf{62.0}_{\hspace{0.05cm}3.7}$ & $\textbf{88.8}_{\hspace{0.05cm}\phantom{1}1.4}$ & $\textbf{76.1}_{\hspace{0.05cm}\phantom{1}7.2}$ & $\textbf{75.7}$\\
\midrule

\parbox[t]{2mm}{\multirow{9}{*}{\rotatebox[origin=c]{90}{\textbf{Mistral-Instruct-7B}}}}
&~  \Vanilla\ $3,4$ & $90.1_{\hspace{0.05cm}2.9}$ & $81.3_{\hspace{0.05cm}2.1}$ & $70.9_{\hspace{0.05cm}7.2}$ & $58.5_{\hspace{0.05cm}4.2}$ & $80.5_{\hspace{0.05cm}1.7}$ & $56.1_{\hspace{0.05cm}5.0}$ & $83.0_{\hspace{0.05cm}5.7}$ & $79.8_{\hspace{0.05cm}\phantom{1}1.4}$ & $75.0$\\
\cmidrule(lr){2-11}
&~  \Vanilla\ $12$ & $91.4_{\hspace{0.05cm}0.9}$ & $81.2_{\hspace{0.05cm}2.2}$ & $67.9_{\hspace{0.05cm}8.7}$ & $57.7_{\hspace{0.05cm}2.8}$ & $79.1_{\hspace{0.05cm}1.6}$ & $57.2_{\hspace{0.05cm}3.6}$ & $85.4_{\hspace{0.05cm}3.6}$ & $77.7_{\hspace{0.05cm}\phantom{1}5.6}$ & $74.7$\\
&~  \PromptS\ $12$ & $90.3_{\hspace{0.05cm}2.5}$ & $81.1_{\hspace{0.05cm}1.9}$ & $68.7_{\hspace{0.05cm}5.8}$ & $57.1_{\hspace{0.05cm}2.7}$ & $79.1_{\hspace{0.05cm}1.6}$ & $56.7_{\hspace{0.05cm}3.2}$ & $84.9_{\hspace{0.05cm}3.0}$ & $79.0_{\hspace{0.05cm}\phantom{1}3.0}$ & $74.6$\\
&~  \CC\ $12$ & $\textbf{91.5}_{\hspace{0.05cm}1.6}$ & $\textbf{81.3}_{\hspace{0.05cm}1.8}$ & $\textbf{75.8}_{\hspace{0.05cm}2.6}$ & $58.3_{\hspace{0.05cm}6.6}$ & $81.0_{\hspace{0.05cm}1.3}$ & $61.9_{\hspace{0.05cm}4.7}$ & $85.4_{\hspace{0.05cm}4.0}$ & $\textbf{79.6}_{\hspace{0.05cm}\phantom{1}1.6}$ & $76.9$\\
&~  \Comp\ $12$ & $89.9_{\hspace{0.05cm}2.7}$ & $80.7_{\hspace{0.05cm}2.7}$ & $75.1_{\hspace{0.05cm}2.9}$ & $\textbf{60.0}_{\hspace{0.05cm}4.9}$ & $\textbf{81.1}_{\hspace{0.05cm}1.3}$ & $\textbf{64.7}_{\hspace{0.05cm}4.6}$ & $\textbf{87.6}_{\hspace{0.05cm}2.1}$ & $79.2_{\hspace{0.05cm}\phantom{1}1.2}$ & $\textbf{77.3}$\\
\cmidrule(lr){2-11}
&~  \Vanilla\ $24$ & $91.2_{\hspace{0.05cm}1.0}$ & $80.8_{\hspace{0.05cm}2.3}$ & $65.3_{\hspace{0.05cm}8.4}$ & $57.4_{\hspace{0.05cm}4.0}$ & $75.6_{\hspace{0.05cm}1.7}$ & $56.6_{\hspace{0.05cm}6.5}$ & $85.8_{\hspace{0.05cm}4.3}$ & $68.8_{\hspace{0.05cm}16.9}$ & $72.7$\\
&~  \PromptS\ $24$ & $90.5_{\hspace{0.05cm}2.6}$ & $\textbf{81.3}_{\hspace{0.05cm}2.0}$ & $68.9_{\hspace{0.05cm}5.6}$ & $57.1_{\hspace{0.05cm}2.1}$ & $79.1_{\hspace{0.05cm}1.7}$ & $57.4_{\hspace{0.05cm}3.1}$ & $86.0_{\hspace{0.05cm}2.1}$ & $78.7_{\hspace{0.05cm}\phantom{1}3.3}$ & $74.9$\\
&~  \CC\ $24$ & $\textbf{91.6}_{\hspace{0.05cm}1.5}$ & $80.9_{\hspace{0.05cm}2.0}$ & $76.1_{\hspace{0.05cm}2.4}$ & $59.5_{\hspace{0.05cm}5.4}$ & $81.2_{\hspace{0.05cm}0.9}$ & $62.7_{\hspace{0.05cm}4.3}$ & $85.9_{\hspace{0.05cm}3.7}$ & $\textbf{80.1}_{\hspace{0.05cm}\phantom{1}1.2}$ & $77.2$\\
&~  \Comp\ $24$ & $90.8_{\hspace{0.05cm}1.8}$ & $80.6_{\hspace{0.05cm}2.1}$ & $\textbf{76.4}_{\hspace{0.05cm}1.7}$ & $\textbf{60.7}_{\hspace{0.05cm}4.4}$ & $\textbf{81.6}_{\hspace{0.05cm}1.0}$ & $\textbf{65.3}_{\hspace{0.05cm}3.4}$ & $\textbf{88.0}_{\hspace{0.05cm}1.8}$ & $79.0_{\hspace{0.05cm}\phantom{1}1.6}$ & $\textbf{77.8}$\\
\midrule

\parbox[t]{2mm}{\multirow{7}{*}{\rotatebox[origin=c]{90}{\textbf{Llama-3-8B}}}}
&~  \Vanilla\ $3,4$ & $91.4_{\hspace{0.05cm}1.7}$ & $79.2_{\hspace{0.05cm}7.2}$ & $74.0_{\hspace{0.05cm}8.0}$ & $58.7_{\hspace{0.05cm}4.7}$ & $76.5_{\hspace{0.05cm}2.2}$ & $59.4_{\hspace{0.05cm}3.7}$ & $84.0_{\hspace{0.05cm}6.6}$ & $87.4_{\phantom{1}\hspace{0.05cm}5.5}$ & $76.3$\\
\cmidrule(lr){2-11}
&~  \Vanilla\ $12$ & $\textbf{92.2}_{\hspace{0.05cm}0.6}$ & $\textbf{79.6}_{\hspace{0.05cm}6.9}$ & $73.1_{\hspace{0.05cm}5.0}$ & $\textbf{63.3}_{\hspace{0.05cm}2.4}$ & $77.5_{\hspace{0.05cm}2.2}$ & $62.7_{\hspace{0.05cm}3.8}$ & $87.7_{\hspace{0.05cm}2.0}$ & $82.1_{\hspace{0.05cm}18.1}$ & $77.3$\\
&~  \CC\ $12$ & $91.1_{\hspace{0.05cm}1.5}$ & $79.2_{\hspace{0.05cm}5.8}$ & $\textbf{77.9}_{\hspace{0.05cm}3.3}$ & $60.5_{\hspace{0.05cm}9.3}$ & $77.3_{\hspace{0.05cm}2.1}$ & $65.2_{\hspace{0.05cm}3.2}$ & $86.4_{\hspace{0.05cm}4.7}$ & $\textbf{87.7}_{\phantom{1}\hspace{0.05cm}4.0}$ & $78.2$\\
&~  \Comp\ $12$ & $90.7_{\hspace{0.05cm}2.0}$ & $78.3_{\hspace{0.05cm}6.7}$ & $77.2_{\hspace{0.05cm}2.9}$ & $61.8_{\hspace{0.05cm}6.4}$ & $\textbf{78.0}_{\hspace{0.05cm}1.8}$ & $\textbf{66.9}_{\hspace{0.05cm}2.4}$ & $\textbf{89.1}_{\hspace{0.05cm}1.0}$ & $87.4_{\phantom{1}\hspace{0.05cm}3.8}$ & $\textbf{78.7}$\\
\cmidrule(lr){2-11}
&~  \Vanilla\ $24$ & $\textbf{92.2}_{\hspace{0.05cm}0.8}$ & $78.2_{\hspace{0.05cm}7.2}$ & $78.0_{\hspace{0.05cm}2.0}$ & $\textbf{63.8}_{\hspace{0.05cm}1.8}$ & $76.2_{\hspace{0.05cm}3.0}$ & $66.4_{\hspace{0.05cm}2.3}$ & $87.9_{\hspace{0.05cm}1.9}$ & $80.6_{\hspace{0.05cm}18.9}$ & $77.9$\\
&~  \CC\ $24$ & $91.7_{\hspace{0.05cm}1.4}$ & $\textbf{80.0}_{\hspace{0.05cm}6.1}$ & $78.3_{\hspace{0.05cm}3.4}$ & $\textbf{63.8}_{\hspace{0.05cm}2.9}$ & $78.1_{\hspace{0.05cm}1.7}$ & $66.0_{\hspace{0.05cm}2.4}$ & $86.7_{\hspace{0.05cm}4.9}$ & $\textbf{87.7}_{\phantom{1}\hspace{0.05cm}3.8}$ & $79.0$\\
&~  \Comp\ $24$ & $91.6_{\hspace{0.05cm}1.7}$ & $79.1_{\hspace{0.05cm}6.9}$ & $\textbf{78.8}_{\hspace{0.05cm}2.7}$ & $63.7_{\hspace{0.05cm}3.3}$ & $\textbf{78.5}_{\hspace{0.05cm}1.4}$ & $\textbf{67.4}_{\hspace{0.05cm}2.6}$ & $\textbf{89.5}_{\hspace{0.05cm}1.1}$ & $87.4_{\phantom{1}\hspace{0.05cm}3.2}$ & $\textbf{79.5}$\\

\bottomrule
\end{tabular}}

\end{center}

\caption{ICL accuracy of 8 classification tasks and the average accuracy (\textbf{Avg.}). The number after a method denotes the number of labeled data used. We run 15 prompts for each task (5 disjoint sets of $K$ labeled data and 3 templates) and report the mean accuracy and standard deviation. \Comp\ achieves the best average accuracy in all setups.}
\label{table:reweight}
\end{table*}
\subsection{Results}
We set $K=\{12, 24\}$.
Table \ref{table:reweight} compares our component reweighting (\Comp) with standard ICL (\Vanilla), prompt selection (\PromptS), and calibration (\CC).
First, we find that simply increasing the number of demonstrations from $4$ to $24$ has limited improvements in ICL accuracy, while the longer prompt greatly increases the inference time.
For example, on Llama-2-7B, \mbox{\Vanilla\ $24$} only improves the average accuracy by $2.6\%$ over \mbox{\Vanilla\ $3, 4$} and the accuracy even decreases on Mistral-Instruct.
Second, \PromptS\ performs the worst in most setups, likely because it is hard to find similar examples from a small pool of $K$ examples, and a bad selection induces majority label biases.
Third, both calibration (\CC) and component reweighting (\Comp) achieve substantially better accuracy than \mbox{\Vanilla\ $3, 4$} with little test-time overhead.
Overall, \Comp\ achieves the best average accuracy in all setups, outperforming \mbox{\Vanilla\ $12$} by $6.0\%, 1.8\%, 2.6\%, 1.4$ on Llama-2-7B, Llama-2-13B, Mistral-Instruct-7B, and Llama-3-8B, respectively, and outperforming \mbox{\Vanilla\ $24$} by $6.0\%, 2.2\%, 5.1\%, 1.6\%$, respectively.
We run one-tailed paired t-tests comparing \Comp\ with \CC\ and find that p-values $<0.05$ in all 8 setups (see Table \ref{table:pvalues}), showing that \Comp\ performs significantly better \CC.

\section{When Do Good Components Emerge?}
\label{sec:dynamics}
\begin{figure}[t]
  \centering
  \includegraphics[width=1.\linewidth]{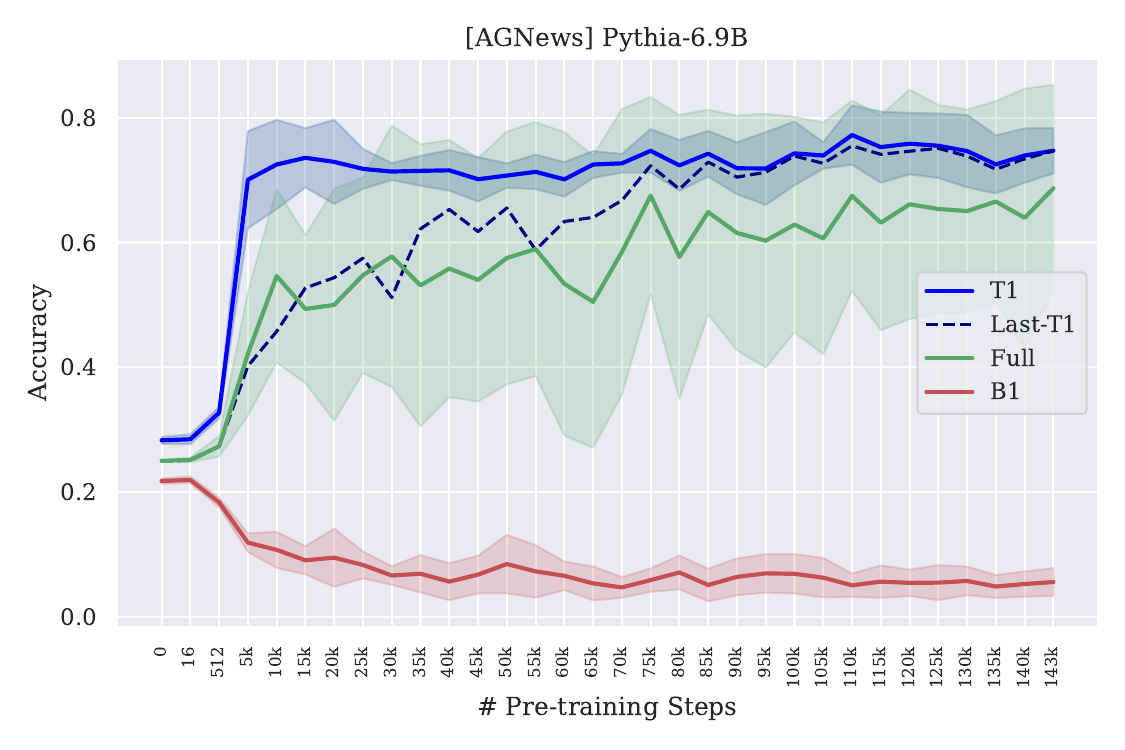}
      \caption{The ICL accuracy of the full model (\textcolor{teal}{green}) fluctuates greatly during pretraining. However, good-performing components (\textcolor{blue}{T1}) emerge in the early steps.}
  \label{fig:dynamics}
\end{figure}

We study the dynamics of components during pretraining by monitoring their accuracies on 32 checkpoints of Pythia-6.9B, uniformly spaced from the first to the last checkpoint.
For each checkpoint, we run 4-shot ICL on AGNews with 3 templates $\times$ 3 sets of demonstrations.
The demonstrations are balanced in labels with randomly shuffled orders.
Figure \ref{fig:dynamics} shows the average accuracy of the 9 runs shaded by the standard deviation.

While the full model (green) fluctuates and has a large variance across prompts, the top-1 components (solid blue) achieve good accuracy at an early step and plateau quickly.
We also backtrack the top-1 components of different prompts at the last checkpoint (dashed blue), monitoring how they perform on average during pretraining.
We observe that they are not the top components at the early stage (there are gaps between the two blue lines before the $75k$ steps), but start to perform steadily well from the middle stage.
Our findings also hold on SST2 and Pythia-1.4B (see Figure \ref{fig:app_dynamics} in the appendix),
suggesting that the model's ability to do a task emerges before it is apparent from the full model on these tasks.\footnote{On the other hand, Pythia models perform poorly on the other tasks over all checkpoints; thus, the training dynamics of the model components on challenging tasks remain unclear.}

\section{Related Work and Discussion}
\paragraph{Improving ICL.}
Prior work shows that ICL performance varies greatly across different choices of demonstrations and templates \cite{zhao2021calibrate,lu-etal-2022-fantastically}.
Specifically, \citet{sclar2024quantifying} and \citet{voronov-etal-2024-mind} find no universally better prompt template that can transfer across tasks and models, implying that it is not easy to explain ICL through prompt engineering.
While several approaches, such as prompt selection \cite{liu-etal-2022-makes, chang-jia-2023-data, fu2023complexitybased}, prompt ensemble \cite{min-etal-2022-noisy,arora2023ask,voronov-etal-2024-mind}, and many-shot ICL \cite{agarwal2024many}, substantially improve accuracy, they treat LLMs like black boxes without understanding the internals. 
Besides, they greatly increase inference time or require a large set of labeled data, which deviates from true few-shot learning \cite{perez2021true}.
In comparison, our paper studies this problem by looking inside the LLMs.
Rather than selecting prompts, we select components in a soft, learnable way.
Our method only requires $\{12,24\}$ examples and has negligible computation overhead over $4$-shot ICL at inference.

\paragraph{Components Interpretation.}
Components interpretation studies the function of different components in a trained model \cite{elhage2021mathematical,shah2024decomposing}, where components could be neurons \cite{radford2017learning,wang-etal-2022-finding-skill,gurnee2023finding}, attention heads \cite{olsson2022context}, and MLPs \cite{geva-etal-2021-transformer}.
To analyze the components, probing \cite{alain2016understanding}, knockout \cite{geva-etal-2023-dissecting,chang-etal-2024-localization,li2023circuit}, patching \cite{wang2023interpretability,goldowsky2023localizing}, and early decoding \cite{nostalgebraist_interpreting_nodate,geva-etal-2022-transformer} are widely used techniques.
For example, \citet{li2024inference} train a linear probe on every attention head to discover the truthful heads inside LLMs.
\citet{NEURIPS2019_2c601ad9} and \citet{voita-etal-2019-analyzing} prune away a large
percentage of attention heads and show that only a few are critical to the performance.
\citet{hendel-etal-2023-context}, \citet{liu2023context}, \citet{merullo-etal-2024-language}, and \citet{todd2024function} view ICL as compressing demonstrations into function vectors, where they remove the demonstrations and modify (\emph{patch}) the LLM activations at certain layers with the function vectors at test time.
Early decoding interprets the investigated components in the textual space by projecting them through the output embedding matrix \cite{geva-etal-2022-transformer}.
Our model decomposition is based on early decoding and we share some similarities with prior work \cite{yu-etal-2023-characterizing,wang2023forbidden}, especially in discovering individual components that perform well on a task. 
Our contributions lie in providing a new view of ICL by decomposition, which reveals the transferability of components across diverse ICL settings. 

\paragraph{Our Method vs. Pruning.}
Our method caches the direct contributions of components to the outputs through the residual stream, i.e., logits = $\sum_j \CompLogit_j$.
Thus, removing $\CompLogit_j$, the direct contribution of the component $j$, does not alter the contributions of the other components.
In comparison, pruning a component changes the activations of the other components in later layers.
In \ref{sec:app_prune}, we show that pruning the good-performing components identified by our method greatly hurts the accuracy, meaning that pruning also defines these components as important \cite{NEURIPS2019_2c601ad9}.
\section{Conclusion}
We introduce a new perspective of ICL via decomposing the model output into the sum of individual contributions of components.
We then identify three types of component characteristics across 3 LLMs and 8 classification tasks.
Our extensive analyses reveal consistency in component accuracy across prompts and suggest the promising direction of improving ICL by selecting components.
To this end, we propose component reweighting, which learns to scale components differently on few-shot examples.
Our method achieves the best average accuracy compared to prior methods.
We hope this work can deepen our grasp of LLMs while motivating more methods for practical use.
\section{Limitations}
Our component reweighting method requires a small set of labeled data $\Ddev$ to train the component weights $w$.
However, we believe it is not unreasonable to have at least $K=12$ labeled examples in total and we compare with baselines using the same $K$ examples.
On the other hand, we do not compare with fine-tuning-based baselines, such as LM-BFF \cite{gao-etal-2021-making}, T-few \cite{liu2022few}, and LoRA \cite{hu2021lora}, because they usually require a larger GPU memory for training and more sophisticated early stopping criteria to prevent overfitting on few-shot examples.
Another limitation is that we only experiment with classification tasks for ease of evaluation.
We leave it for future work to generalize our method to generation tasks by doing decomposition and reweighting at every token during generation.

Despite similarities in model decomposition, the focus of this paper is not circuits in LLMs \cite{wang2023interpretability}.
Thus, we only have limited experiments towards mechanistic understanding of the curious components in \cref{sec:mech} and \ref{sec:app_prune}.
Unlike prior work that uses synthetic tasks to testify whether a head attends to certain tokens \cite{dutta2024think,merullo2024talking}, we work on standard NLP benchmark datasets without obviously correct or incorrect tokens to collect answers, making mechanistic interpretation more challenging.
\section*{Acknowledgements}
We thank Johnny Wei for his valuable suggestions on the paper structure.
We thank Qinyuan Ye, Ryan Wang, Gustavo Lucas Carvalho, Ameya Godbole, Wang Zhu,  Daniel Firebanks-Quevedo, and the anonymous reviewers for their helpful feedback.
This work was funded in part by gifts from Open Philanthropy and Cisco Research,
and was also supported in part by the National Science
Foundation under Grant No. IIS-2403436.
Any opinions, findings, and conclusions or recommendations expressed
in this material are those of the author(s) and do not necessarily
reflect the views of the National Science Foundation.
\bibliography{anthology, custom}

\begin{thebibliography}{75}
\providecommand{\natexlab}[1]{#1}

\bibitem[{Agarwal et~al.(2024)Agarwal, Singh, Zhang, Bohnet, Chan, Anand, Abbas, Nova, Co-Reyes, Chu et~al.}]{agarwal2024many}
Rishabh Agarwal, Avi Singh, Lei~M Zhang, Bernd Bohnet, Stephanie Chan, Ankesh Anand, Zaheer Abbas, Azade Nova, John~D Co-Reyes, Eric Chu, et~al. 2024.
\newblock Many-shot in-context learning.
\newblock \emph{arXiv preprint arXiv:2404.11018}.

\bibitem[{Alain and Bengio(2017)}]{alain2016understanding}
Guillaume Alain and Yoshua Bengio. 2017.
\newblock \href {https://openreview.net/forum?id=ryF7rTqgl} {Understanding intermediate layers using linear classifier probes}.

\bibitem[{Arora et~al.(2023)Arora, Narayan, Chen, Orr, Guha, Bhatia, Chami, and Re}]{arora2023ask}
Simran Arora, Avanika Narayan, Mayee~F Chen, Laurel Orr, Neel Guha, Kush Bhatia, Ines Chami, and Christopher Re. 2023.
\newblock \href {https://openreview.net/forum?id=bhUPJnS2g0X} {Ask me anything: A simple strategy for prompting language models}.
\newblock In \emph{The Eleventh International Conference on Learning Representations}.

\bibitem[{Bach et~al.(2022)Bach, Sanh, Yong, Webson, Raffel, Nayak, Sharma, Kim, Bari, Fevry, Alyafeai, Dey, Santilli, Sun, Ben-david, Xu, Chhablani, Wang, Fries, Al-shaibani, Sharma, Thakker, Almubarak, Tang, Radev, Jiang, and Rush}]{bach-etal-2022-promptsource}
Stephen Bach, Victor Sanh, Zheng~Xin Yong, Albert Webson, Colin Raffel, Nihal~V. Nayak, Abheesht Sharma, Taewoon Kim, M~Saiful Bari, Thibault Fevry, Zaid Alyafeai, Manan Dey, Andrea Santilli, Zhiqing Sun, Srulik Ben-david, Canwen Xu, Gunjan Chhablani, Han Wang, Jason Fries, Maged Al-shaibani, Shanya Sharma, Urmish Thakker, Khalid Almubarak, Xiangru Tang, Dragomir Radev, Mike Tian-jian Jiang, and Alexander Rush. 2022.
\newblock \href {https://doi.org/10.18653/v1/2022.acl-demo.9} {{P}rompt{S}ource: An integrated development environment and repository for natural language prompts}.
\newblock In \emph{Proceedings of the 60th Annual Meeting of the Association for Computational Linguistics: System Demonstrations}, pages 93--104, Dublin, Ireland. Association for Computational Linguistics.

\bibitem[{Bertsch et~al.(2024)Bertsch, Ivgi, Alon, Berant, Gormley, and Neubig}]{bertsch2024context}
Amanda Bertsch, Maor Ivgi, Uri Alon, Jonathan Berant, Matthew~R Gormley, and Graham Neubig. 2024.
\newblock In-context learning with long-context models: An in-depth exploration.
\newblock \emph{arXiv preprint arXiv:2405.00200}.

\bibitem[{Bhagavatula et~al.(2020)Bhagavatula, Bras, Malaviya, Sakaguchi, Holtzman, Rashkin, Downey, tau Yih, and Choi}]{Bhagavatula2020Abductive}
Chandra Bhagavatula, Ronan~Le Bras, Chaitanya Malaviya, Keisuke Sakaguchi, Ari Holtzman, Hannah Rashkin, Doug Downey, Wen tau Yih, and Yejin Choi. 2020.
\newblock \href {https://openreview.net/forum?id=Byg1v1HKDB} {Abductive commonsense reasoning}.
\newblock In \emph{International Conference on Learning Representations}.

\bibitem[{Biderman et~al.(2023)Biderman, Schoelkopf, Anthony, Bradley, O’Brien, Hallahan, Khan, Purohit, Prashanth, Raff et~al.}]{biderman2023pythia}
Stella Biderman, Hailey Schoelkopf, Quentin~Gregory Anthony, Herbie Bradley, Kyle O’Brien, Eric Hallahan, Mohammad~Aflah Khan, Shivanshu Purohit, USVSN~Sai Prashanth, Edward Raff, et~al. 2023.
\newblock Pythia: A suite for analyzing large language models across training and scaling.
\newblock In \emph{International Conference on Machine Learning}, pages 2397--2430. PMLR.

\bibitem[{Brown et~al.(2020)Brown, Mann, Ryder, Subbiah, Kaplan, Dhariwal, Neelakantan, Shyam, Sastry, Askell et~al.}]{brown2020language}
Tom Brown, Benjamin Mann, Nick Ryder, Melanie Subbiah, Jared~D Kaplan, Prafulla Dhariwal, Arvind Neelakantan, Pranav Shyam, Girish Sastry, Amanda Askell, et~al. 2020.
\newblock Language models are few-shot learners.
\newblock \emph{Advances in neural information processing systems}, 33:1877--1901.

\bibitem[{Chang and Jia(2023)}]{chang-jia-2023-data}
Ting-Yun Chang and Robin Jia. 2023.
\newblock \href {https://doi.org/10.18653/v1/2023.acl-long.452} {Data curation alone can stabilize in-context learning}.
\newblock In \emph{Proceedings of the 61st Annual Meeting of the Association for Computational Linguistics (Volume 1: Long Papers)}, pages 8123--8144, Toronto, Canada. Association for Computational Linguistics.

\bibitem[{Chang et~al.(2024)Chang, Thomason, and Jia}]{chang-etal-2024-localization}
Ting-Yun Chang, Jesse Thomason, and Robin Jia. 2024.
\newblock \href {https://doi.org/10.18653/v1/2024.naacl-long.176} {Do localization methods actually localize memorized data in {LLM}s? a tale of two benchmarks}.
\newblock In \emph{Proceedings of the 2024 Conference of the North American Chapter of the Association for Computational Linguistics: Human Language Technologies (Volume 1: Long Papers)}, pages 3190--3211, Mexico City, Mexico. Association for Computational Linguistics.

\bibitem[{Chen et~al.(2023)Chen, Zhao, Yu, McKeown, and He}]{chen-etal-2023-relation}
Yanda Chen, Chen Zhao, Zhou Yu, Kathleen McKeown, and He~He. 2023.
\newblock \href {https://doi.org/10.18653/v1/2023.findings-emnlp.12} {On the relation between sensitivity and accuracy in in-context learning}.
\newblock In \emph{Findings of the Association for Computational Linguistics: EMNLP 2023}, pages 155--167, Singapore. Association for Computational Linguistics.

\bibitem[{Clark et~al.(2019)Clark, Lee, Chang, Kwiatkowski, Collins, and Toutanova}]{clark-etal-2019-boolq}
Christopher Clark, Kenton Lee, Ming-Wei Chang, Tom Kwiatkowski, Michael Collins, and Kristina Toutanova. 2019.
\newblock \href {https://doi.org/10.18653/v1/N19-1300} {{B}ool{Q}: Exploring the surprising difficulty of natural yes/no questions}.
\newblock In \emph{Proceedings of the 2019 Conference of the North {A}merican Chapter of the Association for Computational Linguistics: Human Language Technologies, Volume 1 (Long and Short Papers)}, pages 2924--2936, Minneapolis, Minnesota. Association for Computational Linguistics.

\bibitem[{Clark et~al.(2018)Clark, Cowhey, Etzioni, Khot, Sabharwal, Schoenick, and Tafjord}]{allenaiarc}
Peter Clark, Isaac Cowhey, Oren Etzioni, Tushar Khot, Ashish Sabharwal, Carissa Schoenick, and Oyvind Tafjord. 2018.
\newblock Think you have solved question answering? try arc, the ai2 reasoning challenge.
\newblock \emph{arXiv:1803.05457v1}.

\bibitem[{Dubey et~al.(2024)Dubey, Jauhri, Pandey, Kadian, Al-Dahle, Letman, Mathur, Schelten, Yang, Fan et~al.}]{dubey2024llama}
Abhimanyu Dubey, Abhinav Jauhri, Abhinav Pandey, Abhishek Kadian, Ahmad Al-Dahle, Aiesha Letman, Akhil Mathur, Alan Schelten, Amy Yang, Angela Fan, et~al. 2024.
\newblock The llama 3 herd of models.
\newblock \emph{arXiv preprint arXiv:2407.21783}.

\bibitem[{Dutta et~al.(2024)Dutta, Singh, Chakrabarti, and Chakraborty}]{dutta2024think}
Subhabrata Dutta, Joykirat Singh, Soumen Chakrabarti, and Tanmoy Chakraborty. 2024.
\newblock How to think step-by-step: A mechanistic understanding of chain-of-thought reasoning.
\newblock \emph{arXiv preprint arXiv:2402.18312}.

\bibitem[{Elhage et~al.(2021)Elhage, Nanda, Olsson, Henighan, Joseph, Mann, Askell, Bai, Chen, Conerly et~al.}]{elhage2021mathematical}
Nelson Elhage, Neel Nanda, Catherine Olsson, Tom Henighan, Nicholas Joseph, Ben Mann, Amanda Askell, Yuntao Bai, Anna Chen, Tom Conerly, et~al. 2021.
\newblock A mathematical framework for transformer circuits.
\newblock \emph{Transformer Circuits Thread}, 1:1.

\bibitem[{Fei et~al.(2023)Fei, Hou, Chen, and Bosselut}]{fei-etal-2023-mitigating}
Yu~Fei, Yifan Hou, Zeming Chen, and Antoine Bosselut. 2023.
\newblock \href {https://doi.org/10.18653/v1/2023.acl-long.783} {Mitigating label biases for in-context learning}.
\newblock In \emph{Proceedings of the 61st Annual Meeting of the Association for Computational Linguistics (Volume 1: Long Papers)}, pages 14014--14031, Toronto, Canada. Association for Computational Linguistics.

\bibitem[{Fu et~al.(2023)Fu, Peng, Sabharwal, Clark, and Khot}]{fu2023complexitybased}
Yao Fu, Hao Peng, Ashish Sabharwal, Peter Clark, and Tushar Khot. 2023.
\newblock \href {https://openreview.net/forum?id=yf1icZHC-l9} {Complexity-based prompting for multi-step reasoning}.
\newblock In \emph{The Eleventh International Conference on Learning Representations}.

\bibitem[{Gao et~al.(2021)Gao, Fisch, and Chen}]{gao-etal-2021-making}
Tianyu Gao, Adam Fisch, and Danqi Chen. 2021.
\newblock \href {https://doi.org/10.18653/v1/2021.acl-long.295} {Making pre-trained language models better few-shot learners}.
\newblock In \emph{Proceedings of the 59th Annual Meeting of the Association for Computational Linguistics and the 11th International Joint Conference on Natural Language Processing (Volume 1: Long Papers)}, pages 3816--3830, Online. Association for Computational Linguistics.

\bibitem[{Gardner et~al.(2020)Gardner, Artzi, Basmov, Berant, Bogin, Chen, Dasigi, Dua, Elazar, Gottumukkala, Gupta, Hajishirzi, Ilharco, Khashabi, Lin, Liu, Liu, Mulcaire, Ning, Singh, Smith, Subramanian, Tsarfaty, Wallace, Zhang, and Zhou}]{gardner-etal-2020-evaluating}
Matt Gardner, Yoav Artzi, Victoria Basmov, Jonathan Berant, Ben Bogin, Sihao Chen, Pradeep Dasigi, Dheeru Dua, Yanai Elazar, Ananth Gottumukkala, Nitish Gupta, Hannaneh Hajishirzi, Gabriel Ilharco, Daniel Khashabi, Kevin Lin, Jiangming Liu, Nelson~F. Liu, Phoebe Mulcaire, Qiang Ning, Sameer Singh, Noah~A. Smith, Sanjay Subramanian, Reut Tsarfaty, Eric Wallace, Ally Zhang, and Ben Zhou. 2020.
\newblock \href {https://doi.org/10.18653/v1/2020.findings-emnlp.117} {Evaluating models{'} local decision boundaries via contrast sets}.
\newblock In \emph{Findings of the Association for Computational Linguistics: EMNLP 2020}, pages 1307--1323, Online. Association for Computational Linguistics.

\bibitem[{Geva et~al.(2023)Geva, Bastings, Filippova, and Globerson}]{geva-etal-2023-dissecting}
Mor Geva, Jasmijn Bastings, Katja Filippova, and Amir Globerson. 2023.
\newblock \href {https://doi.org/10.18653/v1/2023.emnlp-main.751} {Dissecting recall of factual associations in auto-regressive language models}.
\newblock In \emph{Proceedings of the 2023 Conference on Empirical Methods in Natural Language Processing}, pages 12216--12235, Singapore. Association for Computational Linguistics.

\bibitem[{Geva et~al.(2022)Geva, Caciularu, Wang, and Goldberg}]{geva-etal-2022-transformer}
Mor Geva, Avi Caciularu, Kevin Wang, and Yoav Goldberg. 2022.
\newblock \href {https://doi.org/10.18653/v1/2022.emnlp-main.3} {Transformer feed-forward layers build predictions by promoting concepts in the vocabulary space}.
\newblock In \emph{Proceedings of the 2022 Conference on Empirical Methods in Natural Language Processing}, pages 30--45, Abu Dhabi, United Arab Emirates. Association for Computational Linguistics.

\bibitem[{Geva et~al.(2021)Geva, Schuster, Berant, and Levy}]{geva-etal-2021-transformer}
Mor Geva, Roei Schuster, Jonathan Berant, and Omer Levy. 2021.
\newblock \href {https://doi.org/10.18653/v1/2021.emnlp-main.446} {Transformer feed-forward layers are key-value memories}.
\newblock In \emph{Proceedings of the 2021 Conference on Empirical Methods in Natural Language Processing}, pages 5484--5495, Online and Punta Cana, Dominican Republic. Association for Computational Linguistics.

\bibitem[{Goldowsky-Dill et~al.(2023)Goldowsky-Dill, MacLeod, Sato, and Arora}]{goldowsky2023localizing}
Nicholas Goldowsky-Dill, Chris MacLeod, Lucas Sato, and Aryaman Arora. 2023.
\newblock Localizing model behavior with path patching.
\newblock \emph{arXiv preprint arXiv:2304.05969}.

\bibitem[{Gurnee et~al.(2023)Gurnee, Nanda, Pauly, Harvey, Troitskii, and Bertsimas}]{gurnee2023finding}
Wes Gurnee, Neel Nanda, Matthew Pauly, Katherine Harvey, Dmitrii Troitskii, and Dimitris Bertsimas. 2023.
\newblock \href {https://openreview.net/forum?id=JYs1R9IMJr} {Finding neurons in a haystack: Case studies with sparse probing}.
\newblock \emph{Transactions on Machine Learning Research}.

\bibitem[{Hendel et~al.(2023)Hendel, Geva, and Globerson}]{hendel-etal-2023-context}
Roee Hendel, Mor Geva, and Amir Globerson. 2023.
\newblock \href {https://doi.org/10.18653/v1/2023.findings-emnlp.624} {In-context learning creates task vectors}.
\newblock In \emph{Findings of the Association for Computational Linguistics: EMNLP 2023}, pages 9318--9333, Singapore. Association for Computational Linguistics.

\bibitem[{Hewitt and Liang(2019)}]{hewitt-liang-2019-designing}
John Hewitt and Percy Liang. 2019.
\newblock \href {https://doi.org/10.18653/v1/D19-1275} {Designing and interpreting probes with control tasks}.
\newblock In \emph{Proceedings of the 2019 Conference on Empirical Methods in Natural Language Processing and the 9th International Joint Conference on Natural Language Processing (EMNLP-IJCNLP)}, pages 2733--2743, Hong Kong, China. Association for Computational Linguistics.

\bibitem[{Holtzman et~al.(2021)Holtzman, West, Shwartz, Choi, and Zettlemoyer}]{holtzman-etal-2021-surface}
Ari Holtzman, Peter West, Vered Shwartz, Yejin Choi, and Luke Zettlemoyer. 2021.
\newblock \href {https://doi.org/10.18653/v1/2021.emnlp-main.564} {Surface form competition: Why the highest probability answer isn{'}t always right}.
\newblock In \emph{Proceedings of the 2021 Conference on Empirical Methods in Natural Language Processing}, pages 7038--7051, Online and Punta Cana, Dominican Republic. Association for Computational Linguistics.

\bibitem[{Hu et~al.(2021)Hu, Wallis, Allen-Zhu, Li, Wang, Wang, Chen et~al.}]{hu2021lora}
Edward~J Hu, Phillip Wallis, Zeyuan Allen-Zhu, Yuanzhi Li, Shean Wang, Lu~Wang, Weizhu Chen, et~al. 2021.
\newblock Lora: Low-rank adaptation of large language models.
\newblock In \emph{International Conference on Learning Representations}.

\bibitem[{Jiang et~al.(2023)Jiang, Sablayrolles, Mensch, Bamford, Chaplot, Casas, Bressand, Lengyel, Lample, Saulnier et~al.}]{jiang2023mistral}
Albert~Q Jiang, Alexandre Sablayrolles, Arthur Mensch, Chris Bamford, Devendra~Singh Chaplot, Diego de~las Casas, Florian Bressand, Gianna Lengyel, Guillaume Lample, Lucile Saulnier, et~al. 2023.
\newblock Mistral 7b.
\newblock \emph{arXiv preprint arXiv:2310.06825}.

\bibitem[{Li et~al.(2024)Li, Patel, Vi{\'e}gas, Pfister, and Wattenberg}]{li2024inference}
Kenneth Li, Oam Patel, Fernanda Vi{\'e}gas, Hanspeter Pfister, and Martin Wattenberg. 2024.
\newblock Inference-time intervention: Eliciting truthful answers from a language model.
\newblock \emph{Advances in Neural Information Processing Systems}, 36.

\bibitem[{Li et~al.(2023)Li, Davies, and Nadeau}]{li2023circuit}
Maximilian Li, Xander Davies, and Max Nadeau. 2023.
\newblock Circuit breaking: Removing model behaviors with targeted ablation.
\newblock \emph{arXiv preprint arXiv:2309.05973}.

\bibitem[{Liu et~al.(2022{\natexlab{a}})Liu, Tam, Muqeeth, Mohta, Huang, Bansal, and Raffel}]{liu2022few}
Haokun Liu, Derek Tam, Mohammed Muqeeth, Jay Mohta, Tenghao Huang, Mohit Bansal, and Colin~A Raffel. 2022{\natexlab{a}}.
\newblock Few-shot parameter-efficient fine-tuning is better and cheaper than in-context learning.
\newblock \emph{Advances in Neural Information Processing Systems}, 35:1950--1965.

\bibitem[{Liu et~al.(2022{\natexlab{b}})Liu, Shen, Zhang, Dolan, Carin, and Chen}]{liu-etal-2022-makes}
Jiachang Liu, Dinghan Shen, Yizhe Zhang, Bill Dolan, Lawrence Carin, and Weizhu Chen. 2022{\natexlab{b}}.
\newblock \href {https://doi.org/10.18653/v1/2022.deelio-1.10} {What makes good in-context examples for {GPT}-3?}
\newblock In \emph{Proceedings of Deep Learning Inside Out (DeeLIO 2022): The 3rd Workshop on Knowledge Extraction and Integration for Deep Learning Architectures}, pages 100--114, Dublin, Ireland and Online. Association for Computational Linguistics.

\bibitem[{Liu et~al.(2023)Liu, Xing, and Zou}]{liu2023context}
Sheng Liu, Lei Xing, and James Zou. 2023.
\newblock In-context vectors: Making in context learning more effective and controllable through latent space steering.
\newblock \emph{arXiv preprint arXiv:2311.06668}.

\bibitem[{Lu et~al.(2022)Lu, Bartolo, Moore, Riedel, and Stenetorp}]{lu-etal-2022-fantastically}
Yao Lu, Max Bartolo, Alastair Moore, Sebastian Riedel, and Pontus Stenetorp. 2022.
\newblock \href {https://doi.org/10.18653/v1/2022.acl-long.556} {Fantastically ordered prompts and where to find them: Overcoming few-shot prompt order sensitivity}.
\newblock In \emph{Proceedings of the 60th Annual Meeting of the Association for Computational Linguistics (Volume 1: Long Papers)}, pages 8086--8098, Dublin, Ireland. Association for Computational Linguistics.

\bibitem[{McDougall et~al.(2023)McDougall, Conmy, Rushing, McGrath, and Nanda}]{mcdougall2023copy}
Callum McDougall, Arthur Conmy, Cody Rushing, Thomas McGrath, and Neel Nanda. 2023.
\newblock Copy suppression: Comprehensively understanding an attention head.
\newblock \emph{arXiv preprint arXiv:2310.04625}.

\bibitem[{Merullo et~al.(2024{\natexlab{a}})Merullo, Eickhoff, and Pavlick}]{merullo-etal-2024-language}
Jack Merullo, Carsten Eickhoff, and Ellie Pavlick. 2024{\natexlab{a}}.
\newblock \href {https://doi.org/10.18653/v1/2024.naacl-long.281} {Language models implement simple {W}ord2{V}ec-style vector arithmetic}.
\newblock In \emph{Proceedings of the 2024 Conference of the North American Chapter of the Association for Computational Linguistics: Human Language Technologies (Volume 1: Long Papers)}, pages 5030--5047, Mexico City, Mexico. Association for Computational Linguistics.

\bibitem[{Merullo et~al.(2024{\natexlab{b}})Merullo, Eickhoff, and Pavlick}]{merullo2024talking}
Jack Merullo, Carsten Eickhoff, and Ellie Pavlick. 2024{\natexlab{b}}.
\newblock Talking heads: Understanding inter-layer communication in transformer language models.
\newblock \emph{arXiv preprint arXiv:2406.09519}.

\bibitem[{Michel et~al.(2019)Michel, Levy, and Neubig}]{NEURIPS2019_2c601ad9}
Paul Michel, Omer Levy, and Graham Neubig. 2019.
\newblock \href {https://proceedings.neurips.cc/paper_files/paper/2019/file/2c601ad9d2ff9bc8b282670cdd54f69f-Paper.pdf} {Are sixteen heads really better than one?}
\newblock In \emph{Advances in Neural Information Processing Systems}, volume~32. Curran Associates, Inc.

\bibitem[{Min et~al.(2022{\natexlab{a}})Min, Lewis, Hajishirzi, and Zettlemoyer}]{min-etal-2022-noisy}
Sewon Min, Mike Lewis, Hannaneh Hajishirzi, and Luke Zettlemoyer. 2022{\natexlab{a}}.
\newblock \href {https://doi.org/10.18653/v1/2022.acl-long.365} {Noisy channel language model prompting for few-shot text classification}.
\newblock In \emph{Proceedings of the 60th Annual Meeting of the Association for Computational Linguistics (Volume 1: Long Papers)}, pages 5316--5330, Dublin, Ireland. Association for Computational Linguistics.

\bibitem[{Min et~al.(2022{\natexlab{b}})Min, Lyu, Holtzman, Artetxe, Lewis, Hajishirzi, and Zettlemoyer}]{min-etal-2022-rethinking}
Sewon Min, Xinxi Lyu, Ari Holtzman, Mikel Artetxe, Mike Lewis, Hannaneh Hajishirzi, and Luke Zettlemoyer. 2022{\natexlab{b}}.
\newblock \href {https://doi.org/10.18653/v1/2022.emnlp-main.759} {Rethinking the role of demonstrations: What makes in-context learning work?}
\newblock In \emph{Proceedings of the 2022 Conference on Empirical Methods in Natural Language Processing}, pages 11048--11064, Abu Dhabi, United Arab Emirates. Association for Computational Linguistics.

\bibitem[{Mishra et~al.(2022)Mishra, Khashabi, Baral, and Hajishirzi}]{mishra-etal-2022-cross}
Swaroop Mishra, Daniel Khashabi, Chitta Baral, and Hannaneh Hajishirzi. 2022.
\newblock \href {https://doi.org/10.18653/v1/2022.acl-long.244} {Cross-task generalization via natural language crowdsourcing instructions}.
\newblock In \emph{Proceedings of the 60th Annual Meeting of the Association for Computational Linguistics (Volume 1: Long Papers)}, pages 3470--3487, Dublin, Ireland. Association for Computational Linguistics.

\bibitem[{nostalgebraist(2020)}]{nostalgebraist_interpreting_nodate}
nostalgebraist. 2020.
\newblock \href {https://www.lesswrong.com/posts/AcKRB8wDpdaN6v6ru/interpreting-gpt-the-logit-lens} {interpreting {GPT}: the logit lens}.

\bibitem[{Olsson et~al.(2022)Olsson, Elhage, Nanda, Joseph, DasSarma, Henighan, Mann, Askell, Bai, Chen et~al.}]{olsson2022context}
Catherine Olsson, Nelson Elhage, Neel Nanda, Nicholas Joseph, Nova DasSarma, Tom Henighan, Ben Mann, Amanda Askell, Yuntao Bai, Anna Chen, et~al. 2022.
\newblock In-context learning and induction heads.
\newblock \emph{arXiv preprint arXiv:2209.11895}.

\bibitem[{Ouyang et~al.(2022)Ouyang, Wu, Jiang, Almeida, Wainwright, Mishkin, Zhang, Agarwal, Slama, Ray et~al.}]{ouyang2022training}
Long Ouyang, Jeffrey Wu, Xu~Jiang, Diogo Almeida, Carroll Wainwright, Pamela Mishkin, Chong Zhang, Sandhini Agarwal, Katarina Slama, Alex Ray, et~al. 2022.
\newblock Training language models to follow instructions with human feedback.
\newblock \emph{Advances in neural information processing systems}, 35:27730--27744.

\bibitem[{Perez et~al.(2021)Perez, Kiela, and Cho}]{perez2021true}
Ethan Perez, Douwe Kiela, and Kyunghyun Cho. 2021.
\newblock \href {https://openreview.net/forum?id=ShnM-rRh4T} {True few-shot learning with language models}.
\newblock In \emph{Advances in Neural Information Processing Systems}.

\bibitem[{Pilehvar and Camacho-Collados(2019)}]{pilehvar-camacho-collados-2019-wic}
Mohammad~Taher Pilehvar and Jose Camacho-Collados. 2019.
\newblock \href {https://doi.org/10.18653/v1/N19-1128} {{W}i{C}: the word-in-context dataset for evaluating context-sensitive meaning representations}.
\newblock In \emph{Proceedings of the 2019 Conference of the North {A}merican Chapter of the Association for Computational Linguistics: Human Language Technologies, Volume 1 (Long and Short Papers)}, pages 1267--1273, Minneapolis, Minnesota. Association for Computational Linguistics.

\bibitem[{Radford et~al.(2017)Radford, Jozefowicz, and Sutskever}]{radford2017learning}
Alec Radford, Rafal Jozefowicz, and Ilya Sutskever. 2017.
\newblock \href {https://arxiv.org/abs/1704.01444} {Learning to generate reviews and discovering sentiment}.
\newblock \emph{ArXiv preprint}, abs/1704.01444.

\bibitem[{Radford et~al.(2019)Radford, Wu, Child, Luan, Amodei, Sutskever et~al.}]{radford2019language}
Alec Radford, Jeffrey Wu, Rewon Child, David Luan, Dario Amodei, Ilya Sutskever, et~al. 2019.
\newblock Language models are unsupervised multitask learners.
\newblock \emph{OpenAI blog}, 1(8):9.

\bibitem[{Ramanujan et~al.(2020)Ramanujan, Wortsman, Kembhavi, Farhadi, and Rastegari}]{ramanujan2020s}
Vivek Ramanujan, Mitchell Wortsman, Aniruddha Kembhavi, Ali Farhadi, and Mohammad Rastegari. 2020.
\newblock What's hidden in a randomly weighted neural network?
\newblock In \emph{Proceedings of the IEEE/CVF conference on computer vision and pattern recognition}, pages 11893--11902.

\bibitem[{Reimers and Gurevych(2019)}]{reimers-gurevych-2019-sentence}
Nils Reimers and Iryna Gurevych. 2019.
\newblock \href {https://doi.org/10.18653/v1/D19-1410} {Sentence-{BERT}: Sentence embeddings using {S}iamese {BERT}-networks}.
\newblock In \emph{Proceedings of the 2019 Conference on Empirical Methods in Natural Language Processing and the 9th International Joint Conference on Natural Language Processing (EMNLP-IJCNLP)}, pages 3982--3992, Hong Kong, China. Association for Computational Linguistics.

\bibitem[{Romanov and Shivade(2018)}]{romanov-shivade-2018-lessons}
Alexey Romanov and Chaitanya Shivade. 2018.
\newblock \href {https://doi.org/10.18653/v1/D18-1187} {Lessons from natural language inference in the clinical domain}.
\newblock In \emph{Proceedings of the 2018 Conference on Empirical Methods in Natural Language Processing}, pages 1586--1596, Brussels, Belgium. Association for Computational Linguistics.

\bibitem[{Rubin et~al.(2022)Rubin, Herzig, and Berant}]{rubin-etal-2022-learning}
Ohad Rubin, Jonathan Herzig, and Jonathan Berant. 2022.
\newblock \href {https://doi.org/10.18653/v1/2022.naacl-main.191} {Learning to retrieve prompts for in-context learning}.
\newblock In \emph{Proceedings of the 2022 Conference of the North American Chapter of the Association for Computational Linguistics: Human Language Technologies}, pages 2655--2671, Seattle, United States. Association for Computational Linguistics.

\bibitem[{Sclar et~al.(2024)Sclar, Choi, Tsvetkov, and Suhr}]{sclar2024quantifying}
Melanie Sclar, Yejin Choi, Yulia Tsvetkov, and Alane Suhr. 2024.
\newblock \href {https://openreview.net/forum?id=RIu5lyNXjT} {Quantifying language models' sensitivity to spurious features in prompt design or: How i learned to start worrying about prompt formatting}.
\newblock In \emph{The Twelfth International Conference on Learning Representations}.

\bibitem[{Shah et~al.(2024)Shah, Ilyas, and Madry}]{shah2024decomposing}
Harshay Shah, Andrew Ilyas, and Aleksander Madry. 2024.
\newblock Decomposing and editing predictions by modeling model computation.
\newblock \emph{arXiv preprint arXiv:2404.11534}.

\bibitem[{Socher et~al.(2013)Socher, Perelygin, Wu, Chuang, Manning, Ng, and Potts}]{socher-etal-2013-recursive}
Richard Socher, Alex Perelygin, Jean Wu, Jason Chuang, Christopher~D. Manning, Andrew Ng, and Christopher Potts. 2013.
\newblock \href {https://aclanthology.org/D13-1170} {Recursive deep models for semantic compositionality over a sentiment treebank}.
\newblock In \emph{Proceedings of the 2013 Conference on Empirical Methods in Natural Language Processing}, pages 1631--1642, Seattle, Washington, USA. Association for Computational Linguistics.

\bibitem[{Todd et~al.(2024)Todd, Li, Sharma, Mueller, Wallace, and Bau}]{todd2024function}
Eric Todd, Millicent Li, Arnab~Sen Sharma, Aaron Mueller, Byron~C Wallace, and David Bau. 2024.
\newblock \href {https://openreview.net/forum?id=AwyxtyMwaG} {Function vectors in large language models}.
\newblock In \emph{The Twelfth International Conference on Learning Representations}.

\bibitem[{Touvron et~al.(2023)Touvron, Martin, Stone, Albert, Almahairi, Babaei, Bashlykov, Batra, Bhargava, Bhosale et~al.}]{touvron2023llama}
Hugo Touvron, Louis Martin, Kevin Stone, Peter Albert, Amjad Almahairi, Yasmine Babaei, Nikolay Bashlykov, Soumya Batra, Prajjwal Bhargava, Shruti Bhosale, et~al. 2023.
\newblock Llama 2: Open foundation and fine-tuned chat models.
\newblock \emph{arXiv preprint arXiv:2307.09288}.

\bibitem[{Vaswani et~al.(2017)Vaswani, Shazeer, Parmar, Uszkoreit, Jones, Gomez, Kaiser, and Polosukhin}]{vaswani2017attention}
Ashish Vaswani, Noam Shazeer, Niki Parmar, Jakob Uszkoreit, Llion Jones, Aidan~N Gomez, \L~ukasz Kaiser, and Illia Polosukhin. 2017.
\newblock \href {https://proceedings.neurips.cc/paper_files/paper/2017/file/3f5ee243547dee91fbd053c1c4a845aa-Paper.pdf} {Attention is all you need}.
\newblock In \emph{Advances in Neural Information Processing Systems}, volume~30.

\bibitem[{Voita et~al.(2019)Voita, Talbot, Moiseev, Sennrich, and Titov}]{voita-etal-2019-analyzing}
Elena Voita, David Talbot, Fedor Moiseev, Rico Sennrich, and Ivan Titov. 2019.
\newblock \href {https://doi.org/10.18653/v1/P19-1580} {Analyzing multi-head self-attention: Specialized heads do the heavy lifting, the rest can be pruned}.
\newblock In \emph{Proceedings of the 57th Annual Meeting of the Association for Computational Linguistics}, pages 5797--5808, Florence, Italy. Association for Computational Linguistics.

\bibitem[{Voronov et~al.(2024)Voronov, Wolf, and Ryabinin}]{voronov-etal-2024-mind}
Anton Voronov, Lena Wolf, and Max Ryabinin. 2024.
\newblock \href {https://doi.org/10.18653/v1/2024.findings-acl.375} {Mind your format: Towards consistent evaluation of in-context learning improvements}.
\newblock In \emph{Findings of the Association for Computational Linguistics ACL 2024}, pages 6287--6310, Bangkok, Thailand and virtual meeting. Association for Computational Linguistics.

\bibitem[{Wang et~al.(2018)Wang, Singh, Michael, Hill, Levy, and Bowman}]{wang-etal-2018-glue}
Alex Wang, Amanpreet Singh, Julian Michael, Felix Hill, Omer Levy, and Samuel Bowman. 2018.
\newblock \href {https://doi.org/10.18653/v1/W18-5446} {{GLUE}: A multi-task benchmark and analysis platform for natural language understanding}.
\newblock In \emph{Proceedings of the 2018 {EMNLP} Workshop {B}lackbox{NLP}: Analyzing and Interpreting Neural Networks for {NLP}}, pages 353--355, Brussels, Belgium. Association for Computational Linguistics.

\bibitem[{Wang et~al.(2023{\natexlab{a}})Wang, Variengien, Conmy, Shlegeris, and Steinhardt}]{wang2023interpretability}
Kevin~Ro Wang, Alexandre Variengien, Arthur Conmy, Buck Shlegeris, and Jacob Steinhardt. 2023{\natexlab{a}}.
\newblock \href {https://openreview.net/forum?id=NpsVSN6o4ul} {Interpretability in the wild: a circuit for indirect object identification in {GPT}-2 small}.
\newblock In \emph{The Eleventh International Conference on Learning Representations}.

\bibitem[{Wang et~al.(2023{\natexlab{b}})Wang, Li, Dai, Chen, Zhou, Meng, Zhou, and Sun}]{wang-etal-2023-label}
Lean Wang, Lei Li, Damai Dai, Deli Chen, Hao Zhou, Fandong Meng, Jie Zhou, and Xu~Sun. 2023{\natexlab{b}}.
\newblock \href {https://doi.org/10.18653/v1/2023.emnlp-main.609} {Label words are anchors: An information flow perspective for understanding in-context learning}.
\newblock In \emph{Proceedings of the 2023 Conference on Empirical Methods in Natural Language Processing}, pages 9840--9855, Singapore. Association for Computational Linguistics.

\bibitem[{Wang et~al.(2023{\natexlab{c}})Wang, Kai, Hariharan, and Shavit}]{wang2023forbidden}
Tony Wang, Miles Kai, Kaivalya Hariharan, and Nir Shavit. 2023{\natexlab{c}}.
\newblock \href {https://openreview.net/forum?id=1kavETZX3Y} {Forbidden facts: An investigation of competing objectives in llama 2}.
\newblock In \emph{Socially Responsible Language Modelling Research}.

\bibitem[{Wang et~al.(2022{\natexlab{a}})Wang, Wen, Zhang, Hou, Liu, and Li}]{wang-etal-2022-finding-skill}
Xiaozhi Wang, Kaiyue Wen, Zhengyan Zhang, Lei Hou, Zhiyuan Liu, and Juanzi Li. 2022{\natexlab{a}}.
\newblock \href {https://doi.org/10.18653/v1/2022.emnlp-main.765} {Finding skill neurons in pre-trained transformer-based language models}.
\newblock In \emph{Proceedings of the 2022 Conference on Empirical Methods in Natural Language Processing}, pages 11132--11152, Abu Dhabi, United Arab Emirates. Association for Computational Linguistics.

\bibitem[{Wang et~al.(2022{\natexlab{b}})Wang, Mishra, Alipoormolabashi, Kordi, Mirzaei, Naik, Ashok, Dhanasekaran, Arunkumar, Stap, Pathak, Karamanolakis, Lai, Purohit, Mondal, Anderson, Kuznia, Doshi, Pal, Patel, Moradshahi, Parmar, Purohit, Varshney, Kaza, Verma, Puri, Karia, Doshi, Sampat, Mishra, Reddy~A, Patro, Dixit, and Shen}]{wang-etal-2022-super}
Yizhong Wang, Swaroop Mishra, Pegah Alipoormolabashi, Yeganeh Kordi, Amirreza Mirzaei, Atharva Naik, Arjun Ashok, Arut~Selvan Dhanasekaran, Anjana Arunkumar, David Stap, Eshaan Pathak, Giannis Karamanolakis, Haizhi Lai, Ishan Purohit, Ishani Mondal, Jacob Anderson, Kirby Kuznia, Krima Doshi, Kuntal~Kumar Pal, Maitreya Patel, Mehrad Moradshahi, Mihir Parmar, Mirali Purohit, Neeraj Varshney, Phani~Rohitha Kaza, Pulkit Verma, Ravsehaj~Singh Puri, Rushang Karia, Savan Doshi, Shailaja~Keyur Sampat, Siddhartha Mishra, Sujan Reddy~A, Sumanta Patro, Tanay Dixit, and Xudong Shen. 2022{\natexlab{b}}.
\newblock \href {https://doi.org/10.18653/v1/2022.emnlp-main.340} {Super-{N}atural{I}nstructions: Generalization via declarative instructions on 1600+ {NLP} tasks}.
\newblock In \emph{Proceedings of the 2022 Conference on Empirical Methods in Natural Language Processing}, pages 5085--5109, Abu Dhabi, United Arab Emirates. Association for Computational Linguistics.

\bibitem[{Williams et~al.(2018)Williams, Nangia, and Bowman}]{williams-etal-2018-broad}
Adina Williams, Nikita Nangia, and Samuel Bowman. 2018.
\newblock \href {https://doi.org/10.18653/v1/N18-1101} {A broad-coverage challenge corpus for sentence understanding through inference}.
\newblock In \emph{Proceedings of the 2018 Conference of the North {A}merican Chapter of the Association for Computational Linguistics: Human Language Technologies, Volume 1 (Long Papers)}, pages 1112--1122, New Orleans, Louisiana. Association for Computational Linguistics.

\bibitem[{Yu et~al.(2023)Yu, Merullo, and Pavlick}]{yu-etal-2023-characterizing}
Qinan Yu, Jack Merullo, and Ellie Pavlick. 2023.
\newblock \href {https://doi.org/10.18653/v1/2023.emnlp-main.615} {Characterizing mechanisms for factual recall in language models}.
\newblock In \emph{Proceedings of the 2023 Conference on Empirical Methods in Natural Language Processing}, pages 9924--9959, Singapore. Association for Computational Linguistics.

\bibitem[{Zhang and Sennrich(2019)}]{zhang2019root}
Biao Zhang and Rico Sennrich. 2019.
\newblock Root mean square layer normalization.
\newblock \emph{Advances in Neural Information Processing Systems}, 32.

\bibitem[{Zhang and Bowman(2018)}]{zhang-bowman-2018-language}
Kelly Zhang and Samuel Bowman. 2018.
\newblock \href {https://doi.org/10.18653/v1/W18-5448} {Language modeling teaches you more than translation does: Lessons learned through auxiliary syntactic task analysis}.
\newblock In \emph{Proceedings of the 2018 {EMNLP} Workshop {B}lackbox{NLP}: Analyzing and Interpreting Neural Networks for {NLP}}, pages 359--361, Brussels, Belgium. Association for Computational Linguistics.

\bibitem[{Zhang et~al.(2015)Zhang, Zhao, and LeCun}]{NIPS2015_250cf8b5}
Xiang Zhang, Junbo Zhao, and Yann LeCun. 2015.
\newblock \href {https://proceedings.neurips.cc/paper_files/paper/2015/file/250cf8b51c773f3f8dc8b4be867a9a02-Paper.pdf} {Character-level convolutional networks for text classification}.
\newblock In \emph{Advances in Neural Information Processing Systems}, volume~28. Curran Associates, Inc.

\bibitem[{Zhao et~al.(2021)Zhao, Wallace, Feng, Klein, and Singh}]{zhao2021calibrate}
Zihao Zhao, Eric Wallace, Shi Feng, Dan Klein, and Sameer Singh. 2021.
\newblock Calibrate before use: Improving few-shot performance of language models.
\newblock In \emph{International conference on machine learning}, pages 12697--12706. PMLR.

\bibitem[{Zhou et~al.(2023)Zhou, Wan, Proleev, Mincu, Chen, Heller, and Roy}]{zhou2023batch}
Han Zhou, Xingchen Wan, Lev Proleev, Diana Mincu, Jilin Chen, Katherine Heller, and Subhrajit Roy. 2023.
\newblock Batch calibration: Rethinking calibration for in-context learning and prompt engineering.
\newblock \emph{arXiv preprint arXiv:2309.17249}.

\end{thebibliography}
\clearpage
\appendix
\section{Appendix}
\label{sec:appendix}

\subsection{Tasks and Templates}
\label{sec:task_details}
\begin{figure*}[t!]
  \centering
  \includegraphics[width=0.8\linewidth]{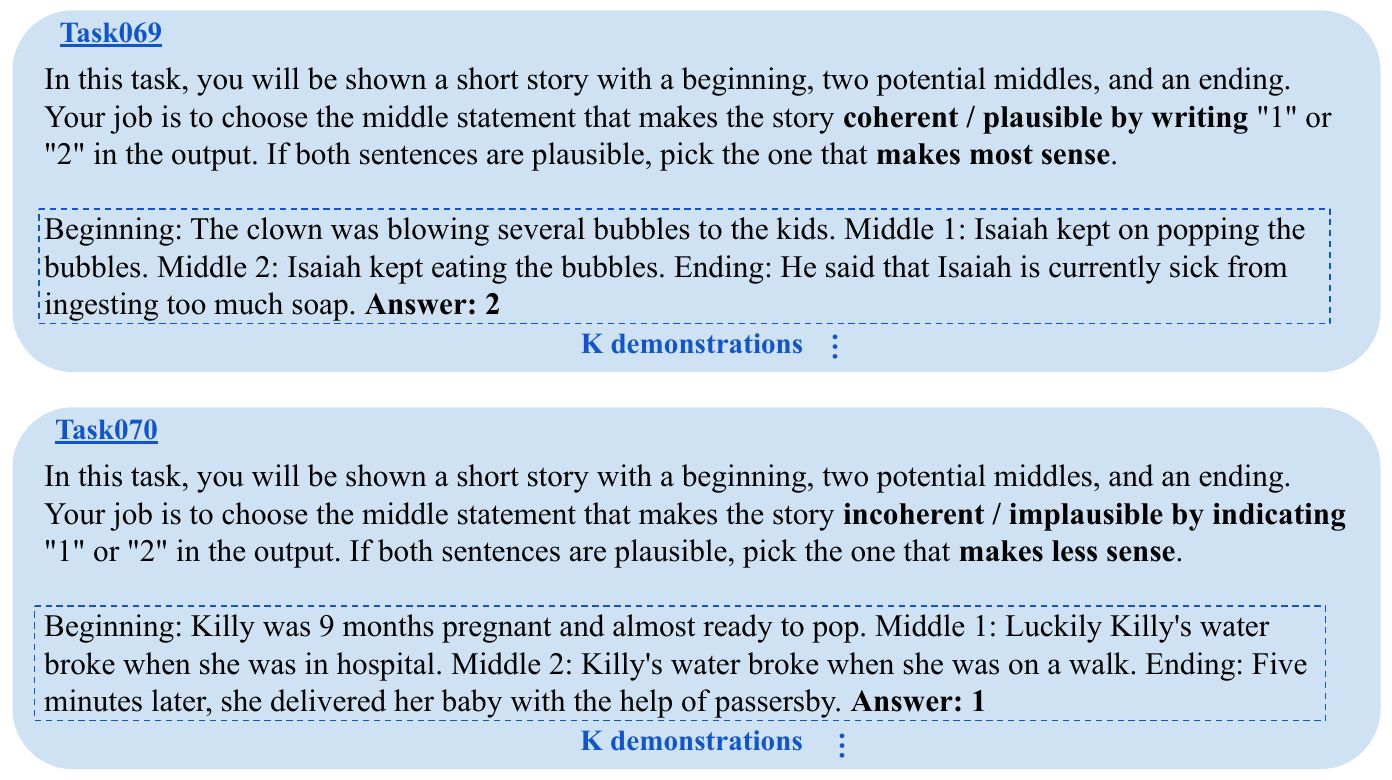}
  \caption{Comparing the prompts of Task069 and Task070. We apply the templates of \citet{sclar2024quantifying} and prepend the task instructions before $K$ demonstrations. We ensure that the two tasks do not have parallel examples to make the transfer experiment (\cref{sec:task69}) challenging.}
  \label{fig:templates_69_70}
\end{figure*}

Table \ref{table:tasks} summarizes the 13 datasets we use in the paper, where we construct balanced test sets by randomly sampling 2000 examples in each task.
We form the prompts by concatenating demonstrations in a randomly shuffled order. To avoid the recency bias \cite{zhao2021calibrate}, we keep shuffling the demonstrations until the last two have different labels.
For minimally conservative templates (\cref{sec:demo_template}), Table \ref{table:appendix_cst} compares the contrast sets we construct on Llama-2-7B.
For our case study on Task069 and Task070, we sample 3 templates from \citet{sclar2024quantifying}.
Figure \ref{fig:templates_69_70} compare the prompts of Task069 and Task070, which consist of an instruction followed by $K$ templated demonstrations.
Originally, the two tasks have $\sim 4\%$ of parallel examples.
To make our task transfer challenging, we discard these overlapped examples.

\subsection{Label-Biased Components}
\label{sec:app_biased}
\begin{figure}[t!]
  \centering
  \includegraphics[width=0.9\linewidth]{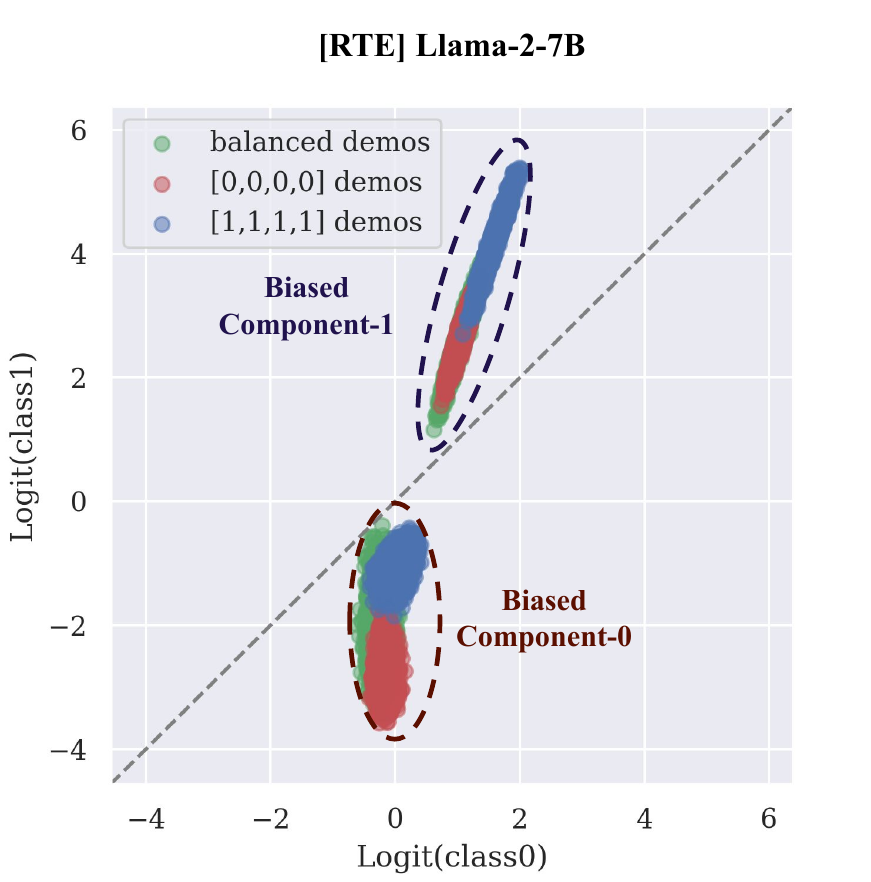}
      \caption{Each dot represents an example in the test set. The two most biased components still insist on predicting the same label on the entire test set regardless of the labels of the demonstrations.}
  \label{fig:biased}
\end{figure}

We say a component is label-biased when it always predicts a certain label on the entire test set \cref{sec:bias}.
In this section, we focus on the most biased components in binary classification tasks, i.e., the two components that have the largest value of $ (\text{logit}_0 - \text{logit}_1)$ and $(\text{logit}_1 - \text{logit}_0)$, respectively, where $\text{logit}_0 \in \RR$ and $\text{logit}_1 \in \RR$ are the LLM output logits on the two classes.
We name these two components as \mbox{Biased Component-0} and Biased Component-1, respectively.
To understand how biased these two components are, we alter the choices of demonstrations and observe their behavior.
Specifically, we consider three settings: demonstrations balanced in labels (green in Figure \ref{fig:biased}), demonstrations of all negative labels ($[0, 0, 0, 0]$; red), and demonstrations of all positive labels ($[1, 1, 1, 1]$; blue).
We fix the template and sample 5 disjoint sets of demonstrations for each setting.
Each dot in Figure \ref{fig:biased} shows the components' prediction on an example, and the x-axis and y-axis correspond to logit$_0$ and logit$_1$, respectively.
A dot below the dashed diagonal line means the prediction on the example is class 0.
We find that both Biased Component-0 and Biased Component-1 still insist on predicting a certain label on all examples, regardless of the labels in the prompts.

\begin{figure}[t!]
  \centering
  \includegraphics[width=0.7\linewidth]{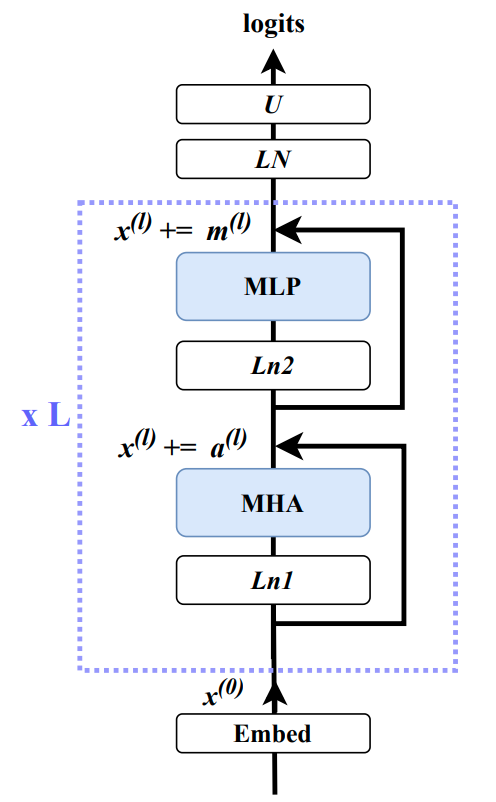}
      \caption{Transformer architecture in GPT2.}
  \label{fig:app_ln}
\end{figure}
\begin{table*}[!t]
\begin{center}
\centering
\resizebox{2.\columnwidth}{!}
{
\begin{tabular}{llccccccccc}
\toprule

& \textbf{SST2} & \textbf{BoolQ} & \textbf{QQP} & \textbf{WiC} & \textbf{RTE} & \textbf{MNLI} & \textbf{AGNews} & \textbf{ARC-Easy} \\

\midrule
\textbf{Llama-2-7B} & $37.8$ & $18.9$ & $43.4$ & $44.2$ & $35.4$ & $28.2$ & $13.4$ & $11.5$ \\
\textbf{Llama-2-13B} & $32.6$ & $18.7$ & $37.2$ & $39.3$ & $32.1$ & $26.0$ & $12.4$ & $13.1$ \\
\textbf{Mistral-Instruct-7B} & $31.9$ & $14.0$ & $32.3$ & $32.4$ & $27.6$ & $20.9$ & $14.5$ & $8.4$ \\
\textbf{Llama-3-8B} & $38.3$ & $21.4$ & $42.4$ & $40.0$ & $36.9$ & $28.1$ & $15.8$ & $15.0$\\
\bottomrule
\end{tabular}}

\end{center}

\caption{We report the average percentage of label-biased components across 15 prompts for each task. A label-biased component always predicts the same label on the entire test set.}
\label{table:label_bias}
\end{table*}

\subsection{LayerNorms}
\label{sec:app_ln}
Figure \ref{fig:app_ln} shows the transformer architecture in GPT2-like models.
Because the layernorms inside each block are before MHA and MLP, known as Pre-LN, Eq. \ref{eq:residual} has already taken $Ln1$ and $Ln2$ into account, and Eq. \ref{eq:flatten} only has the term for the final layernorm, LN$(\cdot)$.

Both Llama-2 and Mistral model families use RMSNorm \cite{zhang2019root}, a layer normalization variant without centering and adding bias terms.
Formally, let $x \in \RR^{d}$ be the input, the root mean square norm LN$(x)$ is:
\begin{align}
    \text{LN}(x) &= \frac{x}{\text{RMS}(x)}\odot \gamma, \\
    \text{RMS}(x) &= \sqrt{\frac{1}{d}\sum_{i=1}^{d} x_i^2},
\end{align}
where $\gamma \in \RR^d$ is the affine transform parameters and $\odot$ denotes element-wise multiplication.

\subsection{Tests of Significance}
We run one-tailed paired t-tests to test whether \Comp\ outperforms \CC\ significantly. In Table \ref{table:reweight}, we have the results of 15 prompts for each task and 8 tasks in total. For each model, we aggregate the 120 accuracy scores of \Comp\ and \CC, respectively, and then calculate the p-values.
Table \ref{table:pvalues} shows that p-values $<0.05$ in 8/8 setups, suggesting that \Comp\ performs significantly better than \CC. 
\begin{table}[!h]
\begin{center}
\centering
\resizebox{1.\columnwidth}{!}
{
\begin{tabular}{lcccc}
\toprule
     &Llama2-7B & Llama2-13B & Mistral-Ins-7B & Llama3-8B\\
\midrule
\textbf{$K=12$} & 0.0010 & 0.0002 & 0.0470 & 0.0198\\
\textbf{$K=24$} & 0.0003 & 0.0001 & 0.0027 & 0.0245 \\

\bottomrule
\end{tabular}}
\end{center}

\caption{The p-values $<0.05$ in all 8 setups (4 LLMs, with $K=\{12, 24\}$ labeled examples), showing that \Comp\ performs significantly better than \CC.}
\label{table:pvalues}
\end{table}

\begin{table}[!t]
\begin{center}
\centering
\resizebox{.85\columnwidth}{!}
{
\begin{tabular}{llcc}
\toprule

&& \textbf{SST2} & \textbf{ARC-Easy} \\

\midrule
\parbox[t]{2mm}{\multirow{3}{*}{\rotatebox[origin=c]{90}{\textbf{Trained}}}}
& \Whole & $75.8_{\hspace{0.05cm}18.1}$ & $57.5_{\hspace{0.05cm}14.4}$ \\
& \OT & $91.7_{\hspace{0.05cm}\phantom{1}0.9}$ & $54.5_{\hspace{0.05cm}10.1}$ \\
& \Comp\ & $90.8_{\hspace{0.05cm}\phantom{1}1.8}$ & $79.0_{\hspace{0.05cm}\phantom{1}1.6}$\\
\midrule
\parbox[t]{2mm}{\multirow{3}{*}{\rotatebox[origin=c]{90}{\textbf{Random}}}}
& \Whole & $49.7_{\hspace{0.05cm}0.7}$ & $25.0_{\hspace{0.05cm}0.5}$ \\
& \OT & $55.2_{\hspace{0.05cm}0.5}$ & $26.7_{\hspace{0.05cm}0.2}$\\
& \Comp\ & $51.4_{\hspace{0.05cm}1.7}$ & $25.0_{\hspace{0.05cm}0.7}$ \\

\bottomrule
\end{tabular}}

\end{center}

\caption{Comparing the ICL accuracy between pretrained (\textbf{Up}) and randomly initialized (\textbf{Down}) Llama-2-7B. The top-1 component (\OT) and \Comp\ perform near random on the untrained model.}
\label{table:app_random}
\end{table}
\subsection{Do Good-Performing Components Exist in Randomly Initialized Models?}
\citet{ramanujan2020s} find that untrained subnetworks can perform on par with a ResNet-34 trained on ImageNet.
Similarly, \citet{zhang-bowman-2018-language,hewitt-liang-2019-designing} show that representations of randomly initialized language models yield a strong baseline for probing tasks.
In this section, we investigate (1) whether good-performing components still exist in a randomly initialized LLM, and (2) how \Comp\ method performs using component activations extracted from the randomly initialized LLM.

We run 4-shot ICL with 15 prompts and report the average accuracy and standard deviation.
For \Comp, we use the same 4 demonstrations and 20 more examples for reweighting.
Table \ref{table:app_random} shows that the best-performing component (\OT) in a randomly initialized Llama-2-7B still performs poorly on SST2 and ARC-Easy.
While \Comp\ has substantial improvement over \Whole\ on the pretrained model, it has no effect on the randomly initialized model.
We conclude that good-performing components do not exist in a randomly initialized LLM and our \mbox{\Comp} method relies on the pretrained component activations to perform well.

\subsection{More Results on Transferability}
\label{sec:full_transfer}
In \cref{sec:transfer}, we study the transferability of components across different choices of demonstrations and templates.
Here, Table \ref{table:iou_full} shows the full results on all LLMs and tasks.
We observe the same findings as Table \ref{table:iou}: component accuracies agree well across randomly sampled demonstrations, but have much weaker agreements across randomly sampled templates.
Because constructing minimally-contrastive templates requires non-trivial manual efforts, we only build contrast sets for 5 tasks on Llama-2-7B (shown in Table \ref{table:iou}), where these tasks have the largest variances across templates.

\subsection{Pruning Good and Bad Components}
\label{sec:app_prune}
Our method studies a component using its cached direct contribution to the output, whereas \citet{NEURIPS2019_2c601ad9} (\emph{pruning}) zeroes out the activations of a component in the forward pass and thus indirectly changes the activations of other components in the upper layers.
They consider a component important if pruning it causes large drops in task performance.
In this section, we investigate the intersection between our method and pruning.

First, we apply our decomposition to identify good and bad-performing components based on their ICL accuracy (3-shot for MNLI, 4-shot for other tasks). 
Second, we run ICL with pruning on Llama-2-7B, using the same 15 prompts in our main experiments for every task.
We prune the top-50 components\footnote{$\sim 5\%$ of the total components} and the bottom-50 components, respectively.
Table \ref{table:prune} compares the results with the full model without pruning.
We find that pruning the top components (T50) greatly hurts the accuracy.
On the contrary, pruning the bottom components (B50) only decreases the average accuracy on SST2 and RTE by $3.5\%$, and even slightly improves the ones on MNLI and AGNews.
These findings may imply that our method and pruning interpret components in similar fashion.
\begin{table}[!h]
\begin{center}
\centering
\resizebox{1.\columnwidth}{!}
{
\begin{tabular}{lcccc}
\toprule

& \textbf{SST2} & \textbf{RTE} & \textbf{MNLI} & \textbf{AGNews}\\

\midrule
\textbf{Full Model} & $\textbf{75.8}_{\hspace{0.05cm}18.1}$ & $\textbf{68.9}_{\hspace{0.05cm} 3.2}$ & $34.4_{\hspace{0.05cm}1.7}$ & $70.0_{\hspace{0.05cm}19.9}$ \\
\textbf{Prune-T50} & $\textcolor{red}{53.4}_{\phantom{1}\hspace{0.05cm}8.8}$ & $\textcolor{red}{57.8}_{\hspace{0.05cm} 6.4}$ & $34.3_{\hspace{0.05cm}3.2}$ & $\textcolor{red}{26.8}_{\phantom{1}\hspace{0.05cm}2.5}$ \\
\textbf{Prune-B50} & $72.3_{\hspace{0.05cm}14.8}$ & $65.4_{\hspace{0.05cm} 5.8}$ & $\textbf{35.7}_{\hspace{0.05cm}4.0}$ & $\textbf{72.6}_{\hspace{0.05cm}15.7}$ \\

\bottomrule
\end{tabular}}

\end{center}

\caption{Comparing the accuracies of the full Llama-2-7B model and pruning the top/bottom 50 components. We run 15 prompts for each task and report the average accuracy and standard deviation. We color the numbers red when there is a large drop in accuracy.}
\label{table:prune}
\end{table}

\subsection{Training Details and Hyperparameters}
\label{sec:app_hyper}
For both \Comp\ and \CC\  methods, we train a linear layer on $\Ddev$ with stochastic gradient descent.
Because we do not have an additional dev set to tune the hyperparameters, we use the same hyperparameters on all the tasks and models and do early stopping based on the loss and accuracy on $\Ddev$.
Specifically, we set learning rate $=0.05$ for both methods and $\lambda = 0.1$ for the L1 regularization term in \Comp.
We run all our ICL experiments on a single RTX A6000 GPU (48G).
Both the component reweighting and calibration training processes can be run on a single i7 CPU within a minute.

\subsection{Models}
We use the model checkpoints on \mbox{\texttt{Hugging Face}}, \texttt{meta-llama/Llama-2-7b-hf}, \texttt{Llama-2-13b-hf}, \texttt{mistralai/Mistral-7B-Instruct-v0.1}, and \texttt{meta-llama/Meta-Llama-3-8B}.

\begin{table*}[!t]
\begin{center}
\centering

{
\begin{tabular}{lcccccccc}
\toprule

& \textbf{SST2} & \textbf{BoolQ} & \textbf{QQP} & \textbf{WiC} & \textbf{RTE} & \textbf{MNLI} & \textbf{AGNews} & \textbf{ARC} \\
\midrule
\textbf{Correlation} & \multicolumn{8}{c}{\emph{\textbf{Llama-2-7B}}} \\
(1) Demo & 0.81 & 0.84 & 0.60 & 0.65 & 0.75 & 0.65 & 0.89 & 0.88 \\
(2) Temp &0.40 & 0.16 & 0.03 & 0.15 & 0.19 & 0.09 & 0.68 & 0.44 \\
\cmidrule(lr){0-0}
\textbf{IoU} \\
(1) Demo & 0.36 & 0.74 & 0.27 & 0.21 & 0.53 & 0.24 & 0.63 & 0.70 \\
(2) Temp & 0.12 & 0.01 & 0.01 & 0.03 & 0.05 & 0.01 & 0.20 & 0.20 \\
\midrule
\textbf{Correlation} & \multicolumn{8}{c}{\emph{\textbf{Llama-2-13B}}}\\
(1) Demo & 0.83 & 0.84 & 0.63 & 0.67 & 0.78 & 0.73 & 0.91 & 0.91 \\
(2) Temp &0.57 & 0.30 & 0.09 & 0.19 & 0.28 & 0.16 & 0.76 & 0.55 \\
\cmidrule(lr){0-0}
\textbf{IoU} \\
(1) Demo & 0.26 & 0.71 & 0.31 & 0.18 & 0.46 & 0.39 & 0.55 & 0.65 \\
(2) Temp & 0.21 & 0.11 & 0.07 & 0.01 & 0.21 & 0.07 & 0.25 & 0.30 \\
\midrule
\textbf{Correlation} & \multicolumn{8}{c}{\emph{\textbf{Mistral-Instruct-7B}}}\\
(1) Demo & 0.88 & 0.91 & 0.72 & 0.75 & 0.87 & 0.82 & 0.92 & 0.97 \\
(2) Temp &0.58 & 0.44 & 0.19 & 0.26 & 0.40 & 0.30 & 0.77 & 0.60 \\
\cmidrule(lr){0-0}
\textbf{IoU} \\
(1) Demo & 0.39 & 0.59 & 0.27 & 0.29 & 0.50 & 0.45 & 0.68 & 0.80 \\
(2) Temp & 0.10 & 0.17 & 0.06 & 0.05 & 0.17 & 0.09 & 0.29 & 0.22 \\
\midrule
\textbf{Correlation} & \multicolumn{8}{c}
{\emph{\textbf{Llama-3-8B}}}\\
(1) Demo & 0.85 & 0.88 & 0.70 & 0.73 & 0.80 & 0.81 & 0.89 & 0.95 \\
(2) Temp & 0.55 & 0.39 & 0.26 & 0.25 & 0.31 & 0.23 & 0.67 & 0.52 \\
\cmidrule(lr){0-0}
\textbf{IoU} \\
(1) Demo & 0.42 & 0.56 & 0.28 & 0.25 & 0.46 & 0.52 & 0.65 & 0.68 \\
(2) Temp & 0.15 & 0.12 & 0.09 & 0.07 & 0.08 & 0.05 & 0.34 & 0.27 \\

\bottomrule
\end{tabular}}

\end{center}

\caption{Full results of the average correlation and IoU between (1) two random sets of demonstrations and (2) two randomly sampled templates.}
\label{table:iou_full}
\end{table*}
\begin{table*}[!t]
\begin{center}
\resizebox{2.\columnwidth}{!}{%
\begin{tabular}{llcc}
\toprule
\textbf{Task} &  \textbf{Templates} & \textbf{Labels} & \textbf{Accuracy}\\
\bottomrule
SST-2 & \begin{tabular}[c]{@{}l@{}}\textbf{T1} Review: \{text\}\textbackslash nDo you think the review is positive or negative? \{label\}\end{tabular} & negative/positive & $50.6 \pm 0.7$\\
& \begin{tabular}[c]{@{}l@{}}\textbf{T2} Review: \{text\}\textcolor{red}{\{space\}}\textbackslash nDo you think the review is positive or negative? \{label\}\end{tabular} & negative/positive & $72.7 \pm 6.1$\\

\midrule
BoolQ & \begin{tabular}[c]{@{}l@{}}\textbf{T1} Based on the following passage, \{question\}? \{passage\}\textcolor{red}{\textbackslash n}Answer: \{label\}\end{tabular} & No/Yes & $ 52.5 \pm 2.0$\\
 & \begin{tabular}[c]{@{}l@{}}\textbf{T2} Based on the following passage, \{question\}? \{passage\} Answer: \{label\}\end{tabular} & No/Yes & $ 66.7 \pm 2.1$\\

\midrule
QQP & \begin{tabular}[c]{@{}l@{}}\textbf{T1} Are the questions "\{sent1\}" and "\{sent2\}" asking the same thing? \{label\}\end{tabular} & \textcolor{red}{no/yes} & $ 54.3 \pm 1.1 $\\
& \begin{tabular}[c]{@{}l@{}}\textbf{T2} Are the questions "\{sent1\}" and "\{sent2\}" asking the same thing? \{label\}\end{tabular} & \textcolor{red}{No/Yes} & $ 68.7 \pm 4.1 $\\

\midrule
 AGNews & \begin{tabular}[c]{@{}l@{}}\textbf{T1} \{text\}\textcolor{red}{\textbackslash n}Is this a piece of news regarding World, Sports, Business, or Technology? \{label\}\end{tabular} & World/Sports/Business/Technology & $ 43.9 \pm 8.7$\\
 & \begin{tabular}[c]{@{}l@{}}\textbf{T2} \{text\} Is this a piece of news regarding World, Sports, Business, or Technology? \{label\}\end{tabular} & World/Sports/Business/Technology & $ 88.5 \pm 0.8$\\
\bottomrule

\end{tabular}
}
\end{center}
\caption{We construct minimally contrastive templates that only differ slightly (colored in red) but yield large differences in 4-shot ICL accuracy on Llama-2-7B. We report the average accuracy and standard deviation across 5 ICL runs with different demonstrations under the same template.}
\label{table:appendix_cst}
\end{table*}

\begin{table*}[!h]
\begin{center}
\centering
{
\begin{tabular}{lcc}
\toprule

\textbf{Dataset} & \textbf{Task} & \textbf{\# Classes} \\
\midrule
\textbf{SST-2} \cite{socher-etal-2013-recursive} & Sentiment Analysis & 2 \\
\textbf{Yelp-polarity} \cite{NIPS2015_250cf8b5} & Sentiment Analysis & 2 \\
\textbf{BoolQ} \cite{clark-etal-2019-boolq} & Yes/No QA & 2 \\
\textbf{BoolQ Contrast Set} \cite{gardner-etal-2020-evaluating} & Yes/No QA & 2 \\
\textbf{QQP} \cite{wang-etal-2018-glue} & Paraphrase Identification & 2\\
\textbf{WiC} \cite{pilehvar-camacho-collados-2019-wic} & Word Sense Disambiguation & 2\\
\textbf{RTE} \cite{wang-etal-2018-glue} & Natural Language Inference & 2\\
\textbf{MNLI} \cite{williams-etal-2018-broad} & Natural Language Inference & 3\\
\textbf{MedNLI} \cite{romanov-shivade-2018-lessons} & NLI in Medical Domain & 3\\
\textbf{AGNews} \cite{NIPS2015_250cf8b5} & Topic Classification & 4\\
\textbf{ARC-Easy} \cite{allenaiarc} & Multiple-Choice QA & 4\\
\textbf{Task069} \cite{mishra-etal-2022-cross,wang-etal-2022-super} & Abductive NLI & 2\\
\textbf{Task070} \cite{mishra-etal-2022-cross,wang-etal-2022-super} & Abductive NLI & 2\\

\bottomrule
\end{tabular}}

\end{center}

\caption{Summary of all the datasets.}
\label{table:tasks}
\end{table*}
\begin{figure*}[t!]
  \centering
  \includegraphics[width=1.\linewidth]{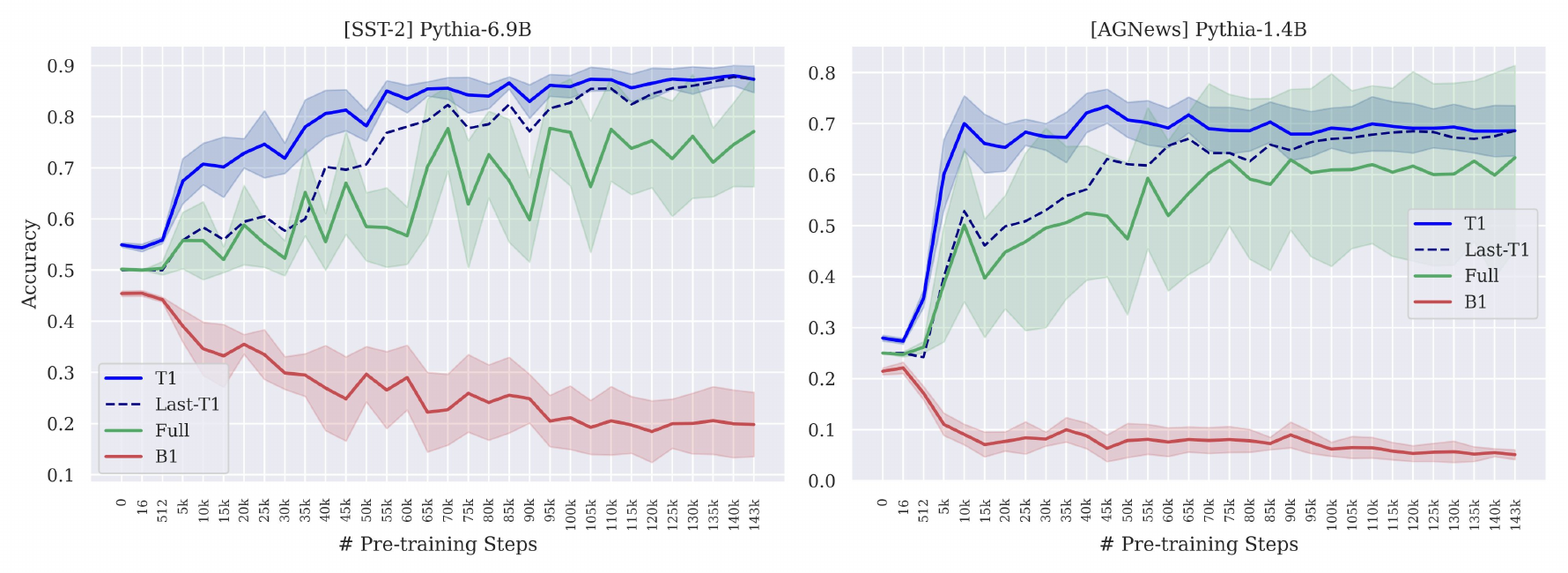}
      \caption{4-shot ICL accuracy on different pretraining checkpoints. We compare the full model (green) with the top-1 (solid blue) and bottom-1 (red) components. The dashed blue line tracks how the top-1 components of the last checkpoint (Last-T1) perform across time.}
  \label{fig:app_dynamics}
\end{figure*}
\begin{figure*}[t!]
  \centering
  \includegraphics[width=0.9\linewidth]{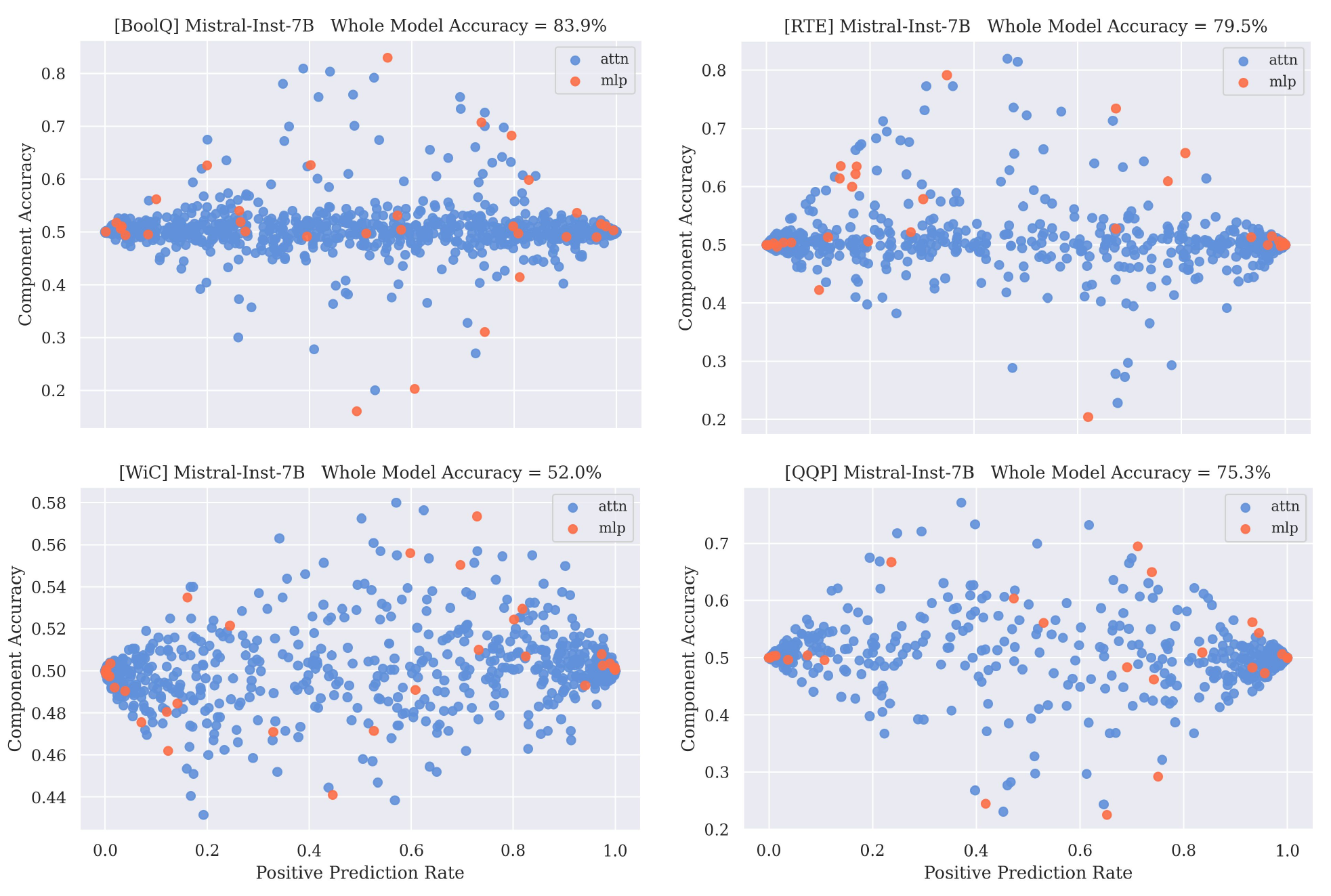}
      \caption{Each dot represents a component (attention head: blue; MLP: orange) under 4-shot ICL on Mistral-Instruct-7B. The x-axis shows how often a component predicts label 1 across the test data of a binary task.}
  \label{fig:app_daimond}
\end{figure*}

\end{document}